\documentclass{article}

     \PassOptionsToPackage{numbers, compress}{natbib}

\usepackage[preprint]{neurips_2024}

\usepackage[utf8]{inputenc} 
\usepackage[T1]{fontenc}    
\usepackage{hyperref}       
\usepackage{url}            
\usepackage{booktabs}       
\usepackage{amsfonts}       
\usepackage{nicefrac}       
\usepackage{microtype}      
\usepackage{xcolor}         
\usepackage{amsmath} 
\usepackage{graphicx}
\usepackage{multirow}

\usepackage{makecell}
\usepackage{tabularray}
\usepackage{colortbl}
\usepackage{bbding}
\usepackage{pifont}
\usepackage{caption}
\usepackage{wrapfig}

\usepackage[defaultcolor=red]{changes} 

\usepackage[capitalize]{cleveref}
\crefname{section}{Sec.}{Secs.}
\Crefname{section}{Section}{Sections}
\Crefname{table}{Table}{Tables}
\crefname{table}{Tab.}{Tabs.}

\title{GeoNLF: Geometry guided Pose-Free \\ Neural LiDAR Fields}

\author{%
  Weiyi Xue, Zehan Zheng, Fan Lu, Haiyun Wei, Guang Chen\thanks{Corresponding author: guangchen@tongji.edu.cn},~Changjun Jiang\\
  Tongji University\\
  \texttt{\{xwy,zhengzehan,lufan,guangchen,haiyunwei,cjjiang\}@tongji.edu.cn} \\
}

\begin{document}

\maketitle


\begin{abstract}
Although recent efforts have extended Neural Radiance Fields (NeRF) into LiDAR point cloud synthesis, the majority of existing works exhibit a strong dependence on precomputed poses. 
However, point cloud registration methods struggle to achieve precise global pose estimation, whereas previous pose-free NeRFs overlook geometric consistency in global reconstruction.
In light of this, we explore the geometric insights of point clouds, which provide explicit registration priors for reconstruction.
Based on this, we propose \textbf{Geo}metry guided \textbf{N}eural \textbf{L}iDAR \textbf{F}ields (GeoNLF), a hybrid framework performing alternately global neural reconstruction and pure geometric pose optimization. Furthermore, NeRFs tend to overfit individual frames and easily get stuck in local minima under sparse-view inputs. 
To tackle this issue, we develop a selective-reweighting strategy and introduce geometric constraints for robust optimization. 
Extensive experiments on NuScenes and KITTI-360 datasets demonstrate the superiority of \textbf{GeoNLF} in both novel view synthesis and multi-view registration of low-frequency large-scale point clouds. 
\end{abstract}
\section{Introduction}
\label{sec:intro}
Neural Radiance Fields (NeRF)~\cite{mildenhall2021nerf} has achieved tremendous achievements in image novel view synthesis (NVS).
Recent studies have extended it to LiDAR point cloud synthesis~\cite{huang2023nfl,tao2023lidarnerf,zhang2023nerflidar,zheng2024lidar4d}, 
mitigating the domain gap to real data and far surpassing traditional methods.
Nevertheless, the majority of existing works exhibit a strong dependence on known precise poses.
In the domain of images, conventional approaches rely on Structure-from-Motion algorithms like COLMAP~\cite{schonberger2016structure} to estimate poses, which are prone to failure with sparse or textureless views. As an alternative, recent works~\cite{bian2023nope,heo2023robust,lin2021barf,park2023camp} such as BARF~\cite{lin2021barf} 
employ bundle-adjusting techniques to achieve high-quality NVS while simultaneously enhancing the precision of pose estimation.

However, the sparse nature of LiDAR point clouds and their inherent absence of texture information distinguish them significantly from images.
Trivial bundle-adjusting techniques from the image domain become less applicable in this context, encountering the following challenges:
(1) Outdoor LiDAR point clouds (\textit{e.g.}, 2Hz, 32-beam LiDAR keyframes in Nuscenes~\cite{caesar2020nuscenes}) exhibit temporal and spatial sparsity. NeRF easily overfits the input views without addressing the geometric inconsistencies caused by inaccurate poses. Consequently, it fails to propagate sufficient gradients for effective pose optimization.
(2) Point clouds lack texture and color information but contain explicit geometric features. However, the photometric-based optimization scheme of NeRFs overlooks these abundant geometric cues within the point cloud, which hinders geometric-based registration.
 
\begin{figure*}[t]
\vspace{-1.3cm}
\centering
  \includegraphics[width=1\textwidth]{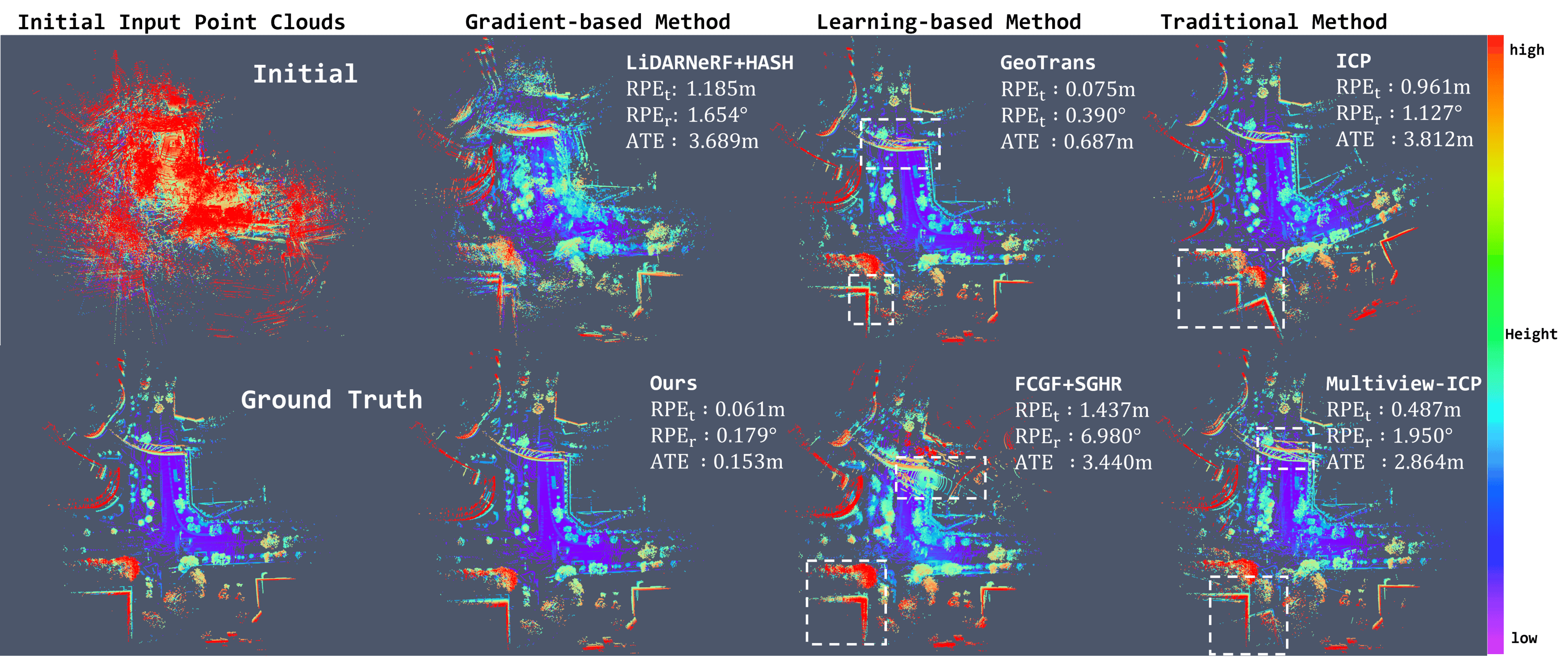}
  \caption{\textbf{Registration results.} Pairwise algorithms such as GeoTrans~\cite{qin2022geometric} and ICP~\cite{besl1992method} suffer from error accumulation and local mismatches. Multi-view methods like SGHR~\cite{wang2023robust} and MICP~\cite{choi2015robust} still manifest outlier poses. Previous gradient-based approaches LiDARNeRF-HASH~\cite{heo2023robust} lack geometric consistency. Our method effectively avoids outlier frames and achieves superior registration accuracy.}
  \label{fig:reg_res}
\vspace{-.2cm}
\end{figure*}

An alternative to achieving pose-free LiDAR-NeRF is to employ point cloud registration (PCR) methods. Nonetheless, as the frequency of point cloud sequences decreases, the inter-frame motion escalates with a reduction in overlap. As presented in \cref{fig:reg_res}, pairwise and multi-view registration approaches may all trap in local optima and suffer from error accumulation, making it challenging to attain globally accurate poses. Hence, integrating local point cloud geometric features for registration with the global optimization of NeRF would be a better synergistic approach. 

Furthermore, as demonstrated in~\cite{bian2023nope,truong2023sparf}, the incorporation of geometric constraints significantly enhances the optimization of both pose and radiance fields. In the image domain, this process involves introducing additional correspondences or depth priors. However, most methods treat them solely as loss terms without fully exploiting them. In contrast, point clouds provide inter-frame correlations (\textit{e.g.}, the closest point) for registration and explicit geometric information for reconstruction, presenting substantial advantages over images. 
To this end, we propose \textbf{GeoNLF}, integrating LiDAR NVS with multi-view PCR
for large-scale and low-frequency point clouds. Specifically, 
to address the 
suboptimality of global optimization and guide NeRF in the early pose optimization stage to avoid local minima,
we regulate NeRF
with a pure geometric optimizer. 
This module constructs a graph for multi-view point clouds and optimizes poses through graph-based loss.
In furtherance of mitigating overfitting, 
we devised a selective-reweighting technique involving filtering out frames with outlier poses, thereby lessening their deleterious impacts throughout the optimization process.
Additionally, to fully leverage the geometric attributes of point clouds, we introduced geometric constraints for point cloud modality 
rather than relying solely on the range map for supervision.
Furthermore, our approach has demonstrated excellent performance in large-scale scenarios with sparse point cloud sequences at 2Hz, spanning hundreds of meters.
To summarize, our main contributions are as follows:

(1) We propose \textbf{GeoNLF}, a novel framework for simultaneous large-scale multi-view PCR and LiDAR NVS.
By exploiting geometric clues inside point clouds, \textbf{GeoNLF} couples geometric optimizer
with neural reconstruction in the pose-free paradigm.  
(2) We introduce a selective-reweighting method to effectively alleviate overfitting, which presents excellent robustness across various scenarios.
(3) Comprehensive experiments demonstrate \textbf{GeoNLF} outperforms state-of-the-art methods by a large margin on challenging large-scale and low-frequency point cloud sequences.


\section{Background and Related Work}
\label{sec:related work}

\textbf{Neural Radiance Fields.} NeRF~\cite{mildenhall2021nerf}and related works have achieved remarkable achievements in NVS.
Various neural representations~\cite{barron2021mip,chan2022efficient,chen2022tensorf,hu2023tri,muller2022instant}, such as hash grids~\cite{muller2022instant}, triplanes~\cite{chan2022efficient,hu2023tri} and diverse techniques~\cite{niemeyer2022regnerf,oechsle2021unisurf,wang2021neus,zhang2022ray} have been proposed to enhance NeRF's performance.
Due to the lack of geometric information in images, some methods~\cite{deng2022depth,roessle2022dense,wei2021nerfingmvs,yu2022monosdf} introduce depth prior or point clouds as auxiliary data to ensure multi-view geometric consistency. However, the geometric information and consistency encapsulated in point clouds are still not fully explored and utilized.

\textbf{Novel View Synthesis for LiDAR}. Traditional simulators~\cite{dosovitskiy2017carla,koenig2004design,shah2018airsim} and explicit reconstruct-then-simulate~\cite{guillard2022learning,li2023pcgen,manivasagam2020lidarsim} method exhibit large domain gap compared to real-world data. 
Very recently, a few studies have pioneered in NVS of LiDAR point clouds based on NeRF, surpassing traditional simulation methods. Among them, NeRF-LiDAR~\cite{zhang2024nerf} and UniSim~\cite{yang2023unisim} require both RGB images as inputs. 
LiDAR-NeRF~\cite{tao2023lidarnerf} and NFL~\cite{huang2023nfl} firstly proposed the differentiable LiDAR NVS framework, and
LiDAR4D~\cite{zheng2024lidar4d}
further extended to dynamic scenes. However, most of these approaches still require a pre-computed pose of each point cloud frame and lack attention to geometric properties.

\textbf{Point Cloud Registration}. 
ICP~\cite{besl1992method} and its variants~\cite{ramalingam2013theory,rusinkiewicz2001efficient,pomerleau2013comparing} are the most classic methods for registration, which rely on good initial conditions but are prone to falling into local optima. 
Learning-based method can be categorized into two schemes, i.e., end-to-end registration~\cite{yuan2020deepgmr,li2020iterative,huang2020feature,wang2019deep,aoki2019pointnetlk} and feature matching-based registration such as FCGF~\cite{choy2019fully}. Recently, the specialized outdoor point cloud registration methods HRegNet~\cite{lu2021hregnet} and HDMNet~\cite{xue2024hdmnet} have achieved excellent results. GeoTransformer~\cite{qin2022geometric} has achieved state-of-the-art in both indoor and outdoor point cloud registration. 
However, learning-based methods are data-driven and limited to specific datasets with ground truth poses, which requires costly pretraining and suffers from poor generalization.

Multiview methods are mostly designed for indoor scenes. Apart from Multiview-ICP~\cite{choi2015robust,birdal2017cad,maset2017practical}, modern methods~\cite{arie2012global,birdal2019probabilistic,tejus2023rotation,jin2024multiway} take global cycle consistency to optimize poses starting from an initial set of pairwise maps. Recent developments~\cite{gojcic2020learning,wang2023robust,arrigoni2016spectral} such as SGHR~\cite{wang2023robust} employ an iteratively reweighted least-squares (IRLS) scheme to adaptively downweight noisy pairwise estimates. However, their registration accuracy fundamentally depends on pairwise registration. The issues of pairwise methods for NVS still persist. 

\textbf{Bundle-Adjusting NeRF}. iNeRF~\cite{yen2021inerf} and subsequent works~\cite{lin2023parallel,deng2023nerf} demonstrated the ability of a trained NeRF to estimate novel view image poses through gradient descent. NeRFmm~\cite{wang2021nerf} and SCNeRF~\cite{song2023sc} extend the method to intrinsic parameter estimation. 
BARF~\cite{lin2021barf} uses a coarse-to-fine reconstruction scheme in gradually learning positional encodings, demonstrating notable efficacy. Subsequent work HASH~\cite{heo2023robust} adapts this approach on iNGP~\cite{muller2022instant} through a weighted schedule of different resolution levels, further boosting performance.
Besides, some studies have extended BARF to address more challenging scenarios, such as sparse input~\cite{truong2023sparf}, dynamic scenes~\cite{liu2023robust} and generalizable NeRF~\cite{chen2023dbarf}. And ~\cite{bian2023nope,truong2023sparf} uses monocular depth or correspondences priors for scene constraints, significantly enhancing the optimization of both pose and radiance fields. However, the aforementioned methods cannot be directly applied to point clouds or experience dramatic performance degradation when transferring. In contrast, our work is the first to introduce bundle-adjusting NeRF into LiDAR NVS task and achieve excellent results in challenging outdoor scenarios.

\section{Methodology}
\label{sec:method}
We firstly introduce the pose-free Neural LiDAR Fields and the problem formulation of pose-free LiDAR-NVS. Following this, a detailed description of our proposed \textbf{GeoNLF} framework is provided. 

\textbf{Pose-Free NeRF and Neural LiDAR Fields.} 
NeRF represents a 3D scene implicitly by encoding the density $\sigma$ and color $\boldsymbol{c}$ of the scene using an implicit neural function $F_{\Theta}(\boldsymbol{x},\boldsymbol{d})$, where $\boldsymbol{x}$ is the 3D coordinates and $\boldsymbol{d}$ is the view direction. When synthesizing novel views, NeRF employs volume rendering techniques to accumulate densities and colors along sampled rays. 
While NeRF requires precise camera parameters, pose-free NeRF only uses images $\mathcal{I}=\{I_i | i=0,1...,N-1\}$ and treats camera parameters $P=\{P_s | s=0,1...N-1\}$ as learnable parameters similar to $\Theta$. Hence, the simultaneous update via gradient descent of P and $\Theta$ can be achieved by minimizing the error $\mathcal{L} = \sum_{i=0}^N \|\hat{I}_i-I_i\|^2_2$ between the rendered and ground truth image $\hat{I},I$:
\begin{align}
    \Theta^*, \mathcal{P}^*=\arg \min _{\Theta,\mathcal{P}} \mathcal{L}(\hat{\mathcal{I}},\hat{\mathcal{P}} \mid \mathcal{I})
    \label{eq:volume_rendering}
\end{align}
Following ~\cite{zheng2024lidar4d,tao2023lidarnerf}, we
project the LiDAR point clouds into range image, 
then cast a ray with a direction $\boldsymbol{d}$ determined by the azimuth angle $\theta$ and elevation angle $\phi$ under the polar coordinate system:
$\mathbf{d} = (\cos\theta \cos \phi, \ \sin \theta \sin \phi, \ \cos \phi)^T$.
Like pose-free NeRF, our pose-free Neural LiDAR Fields treats LiDAR poses as learnable parameters and applies neural function $F_\Theta$ to obtain a radiance depth $z$ and a volume density value $\sigma$. Subsequently, volume rendering techniques are employed to derive the pixel depth value $\hat{\mathcal{D}}$:
\begin{small}
\begin{equation} \label{eq_render_depth}
    \hat{\mathcal{D}}(\mathbf{r}) = \sum_{i=1}^{N}T_i \left( 1 - e^{-\sigma_i\delta_i} \right) z_i, \quad 
    T_i = \exp ( -\sum_{j=1}^{i-1}\sigma_j\delta_j )
\end{equation} 
\end{small}
where $\delta$ refers to the distance between samples. We predict the intensity $\mathcal{S}$ and ray-drop probability $\mathcal{R}$ separately in the same way. 
Besides, our pose-free Neural LiDAR Fields adopted the Hybrid Planar-Grid representation from~\cite{zheng2024lidar4d} for positional encoding $\gamma(x,y,z) =\mathbf{f}_\mathrm{planar} \oplus \mathbf{f}_\mathrm{hash}$. 
\begin{equation} \label{T}
    \mathbf{f}_\mathrm{planar}\!=\!\prod_{i=1}^{3}\mathrm{Bilinear}(\mathcal{V},\boldsymbol{x}),
    \mathcal{V} \in \mathbb{R}^{3\!\times\!M \!\times\!M\!\times\!C}, \ 
    \mathbf{f}_\mathrm{hash}\!=\!\mathrm{TriLinear}(\mathcal{H},\boldsymbol{x}),
    \mathcal{G} \in \mathbb{R}^{M\!\times\! M\!\times\!M\!\times\!C}
\end{equation} 
where $\boldsymbol{x}$ is the 3D point, $\mathcal{V},\mathcal{H}$ store the grid features with $M$ spatial resolution and $C$ feature channels. This encoding method benefits the representation of large-scale scenes\cite{zheng2024lidar4d}.

\textbf{Problem Formulation}.
In the context of large-scale outdoor driving scenarios, the collected LiDAR point cloud sequence $\mathcal{Q}=\{Q_s | s=0,1,...,N\!-\!1\}$ serves as inputs with a low sampling frequency. The goal of GeoNLF is to reconstruct this scene as a continuous implicit representation based on neural fields, jointly recovering the LiDAR poses $P=\{P_s | s=0,1,...,N\!-\!1\}$ which can align all point clouds $\mathcal{Q}$ globally.

\subsection{Overview of GeoNLF Framework}
\label{3.1}
\begin{figure*}[t]
\vspace{-1.2cm}
\centering
  \includegraphics[width=0.99\textwidth]{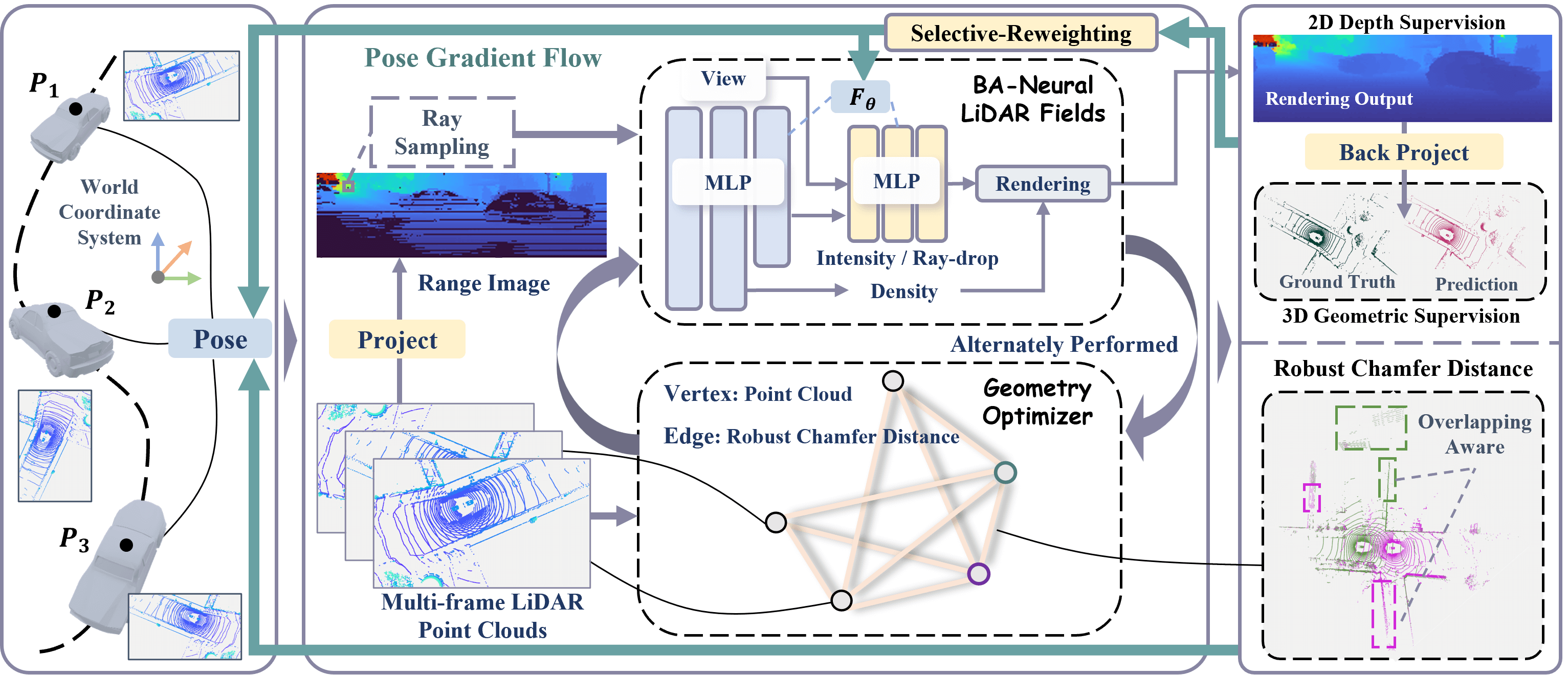}
  \caption{\textbf{Overview of our proposed GeoNLF.} We alternatively execute global optimization of bundle-adjusting neural LiDAR fields and graph-based pure geometric optimization. By integrating selective-reweighting strategy and explicit geometric constraints derived from point clouds, GeoNLF implements outlier-aware and geometry-aware mechanisms.}
  \label{fig:overview}
\vspace{-.2cm}
\end{figure*} 
In contrast to prior pose-free NeRF methods, our pipeline employs a hybrid approach to optimize poses. As shown in \cref{fig:overview}, the framework can be divided into two alternately executed parts: global optimization of bundle-adjusting neural LiDAR fields (\cref{3.2}) and graph-based pure geometric optimization (\cref{3.3}) with the proposed Geo-optimizer. In the first part, we adopt a coarse-to-fine training strategy~\cite{lin2021barf} and extend it to the Hybrid Planar-Grid encoding~\cite{zheng2024lidar4d}. 
In the second part, inspired by multi-view point cloud registration, we construct a graph between multiple frame point clouds and propose a graph-based loss. The graph enables us to achieve pure geometric optimization, which encompasses both inter-frame and global optimization. Furthermore, we integrate the selective-reweighting strategy (\cref{3.4}) into the global optimization.
This encourages the gradient of outliers to propagate towards pose correction while lowering the magnitude transmitted to the radiance fields, thus mitigating the adverse effects of outliers during reconstruction. 
To ensure geometry-aware results, we additionally incorporate explicit geometric constraints derived from point clouds in \cref{3.5}.
\subsection{Bundle-Adjusting Neural LiDAR Fields for Global Optimization}
\label{3.2}
In the stage of global optimization, we optimize Neural LiDAR Fields while simultaneously backpropagating gradients to the pose of each frame. By optimizing our geometry-constrained loss, which will be detailed in \cref{3.5}, the pose is individually optimized to achieve global alignment.

\textbf{LiDAR Pose Representation}.
In previous pose-free NeRF methods, poses are often modeled by $\boldsymbol{T}=[\boldsymbol{R}\mid \boldsymbol{t}] \in SE(3)$ with a rotation $\boldsymbol{R} \in SO(3)$ and a translation $\boldsymbol{t} \in \mathbb{R}^3$. Pose updates are computed in the special Euclidean Lie algebra $\mathfrak{se}(3) = \{ \boldsymbol{\xi}=\begin{bmatrix} \boldsymbol{\rho} \\ \boldsymbol{\phi} \\ \end{bmatrix},\boldsymbol{\rho} \in \mathbb{R}^3,\boldsymbol{\phi} \in \mathfrak{so}(3)\}$ by $\boldsymbol{\xi}'=\boldsymbol{\xi}+\Delta \boldsymbol{\xi}$, followed by the exponential map to obtain the transformation matrix $\boldsymbol{T}$:
\begin{align}
    \boldsymbol{T} = \mathrm{exp}(\boldsymbol{\xi^{\wedge}})=
    \sum_{n=0}^{\infty} \frac{1}{n !} (\boldsymbol{\xi^{\wedge}})^n=
    \left[\begin{array}{cc}
    \sum_{n=0}^{\infty} \frac{1}{n !}\left(\boldsymbol{\phi^{\wedge}}\right)^n & \sum_{n=0}^{\infty} \frac{1}{(n+1) !}\left(\boldsymbol{\phi}^{\wedge}\right)^n \boldsymbol{\rho} \\
    \mathbf{0}^T & 1
    \end{array}\right]
    \label{se3-SE3}
\end{align}
where $\xi^{\wedge}=\left[\begin{array}{cc} \boldsymbol{\phi}^{\wedge} & \boldsymbol{\rho} \\ \mathbf{0}^T & 0 \end{array}\right]$ and $\phi^{\wedge}$ is the antisymmetric matrix of $\boldsymbol{\phi}$. Given a rotation vector $\boldsymbol{\phi} \in \mathfrak{so}(3)$, rotation matrix $\boldsymbol{R}$ can be obtained through the exponential map $\boldsymbol{R}=\mathrm{exp}(\boldsymbol{\phi}^{\wedge})=\sum_{n=0}^{\infty} \frac{1}{n !}\left(\phi^{\wedge}\right)^n$. Simultaneously, we denote $\sum_{n=0}^{\infty} \frac{1}{(n+1) !}\left(\phi^{\wedge}\right)^n$ as $\boldsymbol{J}$. Then \cref{se3-SE3} can be rewritten as:
\begin{align}
    \boldsymbol{T} =    
    \left[\begin{array}{cc}
    \boldsymbol{R} & \textcolor{red}{\boldsymbol{J}}\boldsymbol{\rho} \\
    \mathbf{0}^T & 1
    \end{array}\right]
    \label{se3-SE3-2}
\end{align}
Consequently, due to the coupling between $\boldsymbol{R}=\sum_{n=0}^{\infty} \frac{1}{n !}\left(\phi^{\wedge}\right)^n$ and $J=\sum_{n=0}^{\infty} \frac{1}{(n+1) !}\left(\phi^{\wedge}\right)^n$, the translation updates are influenced by rotation.
Incorporating momentum may lead to non-intuitive optimization trajectories~\cite{lin2023parallel}. Therefore, we omit the coefficient $\boldsymbol{J}$ from the translation term. This approach enables updating the translation of the origin and the rotation around the origin independently. The detailed derivation is provided in \cref{apd:1}.

\textbf{Coarse-to-Fine Positional Encoding}.
BARF\cite{lin2021barf}/HASH~\cite{heo2023robust} propose to gradually activate high-frequency/high-resolution components within positional encoding. We further apply this approach to multi-scale planar and hash encoding~\cite{zheng2024lidar4d} and found it also yields benefits in our large-scale scenarios. For the detailed formulation, we direct readers to reference \cite{heo2023robust}.

\subsection{Graph-based Pure Geometric Optimization}
\label{3.3}
ICP~\cite{besl1992method} is a classic method for registration based on inter-frame geometric correlations. The essence of ICP lies in searching for the closest point as correspondence in another frame's point at each iteration, followed by using Singular Value Decomposition (SVD) to solve \cref{eq: icp}, then iteratively refining the solution. Nonetheless, ICP frequently converges to local optima (\cref{fig:reg_res}). 
In contrast, NeRF optimizes pose globally through the implicit radiance fields. However, it lacks geometric constraints and overlooks the strong geometric information inherent in the point cloud, leading to poor geometric consistency.
As a consequence, both ICP and NeRF acting individually tend to converge to local optima. Our goal is to employ a hybrid method, utilizing NeRF for global pose optimization and integrating geometric information as an auxiliary support. 

Drawing inspiration from ICP~\cite{besl1992method}, we recognize that minimizing the Chamfer Distance (CD) is in line with the optimization objective of each step in ICP algorithm, as demonstrated in \cref{eq: CD}:
\begin{equation}
\setlength\belowdisplayskip{0.05cm}
\label{eq: icp}
\begin{split}
\mathbf{T}^* = \min _{\mathbf{T}} 
\sum_{\mathbf{p}_{i} \in \mathcal{P}}
\min_{\mathbf{q}_i \in \mathcal{Q}}
\left\| \mathbf{T} \cdot \mathbf{p}_{i}-\mathbf{q}_{i} \right\|_2^2 
\end{split}
\end{equation}
\begin{small}
\begin{equation}
\label{eq: CD}
\begin{split}
    \mathcal{L}_{(P,Q)} = 
    \sum_{\mathbf{p}_{i} \in \mathcal{P}} 
    w_i \min_{\mathbf{q}_i \in \mathcal{Q}} \left\| \mathbf{T_p}  p_i   -\mathbf{T_q}  q_i \right\|^{2}_{2}
    +\sum_{\mathbf{q}_{i} \in \mathcal{Q}} 
    w_i \min_{\mathbf{p}_i \in \mathcal{P}} \left\|\ \mathbf{T_q}  q_i - \mathbf{T_p}  p_i \right\|^{2}_{2}
\end{split}
\end{equation}
\end{small}
where $q,p$ in point cloud $\mathbf{Q},\mathbf{P}$ are homogeneous coordinates. $\mathbf{T}_P,\mathbf{T}_Q$ represent the transformation matrix to the world coordinate system. However, minimizing the original CD does not necessarily indicate improved accuracy due to the non-overlapping regions between point clouds. 
To alleviate this negative impact, we weight each correspondence based on \cref{eq: weight}, whereas $w_i$ in the original CD is normalized by the $\frac{1}{N}$, N is the number of points.
\begin{equation}
\label{eq: weight}
    w_{i} = \frac{\mathrm{exp}(t/d_i)}{\sum_{i=1}^N \mathrm{exp}(t/d_i)}, \quad
    t=\mathrm{scheduler}(t_0),\quad
    d_{clipped}=\mathrm{max}(\mathrm{voxelsize},d)
\end{equation}
where $t$ is the temperature to sharpen the distribution of the $d_i$. The distance $d$ is clipped to the size of the downsampled voxel grid. This soft assignment can be considered as an approximately derivable version of weighted averaging. \cref{eq: CD} will degenerate to the original CD when $t \rightarrow 0$, degrade to considering only correspondences with the minimum distance when $t \rightarrow \infty$. Considering the distance lacks practical significance in initial optimization, the scheduler is set as a linear or exponential function to vary $t$ from 0 to 0.5 as the optimization progresses. 
\begin{figure*}[t]
\vspace{-1.3cm}
\centering
  \includegraphics[width=1\textwidth]{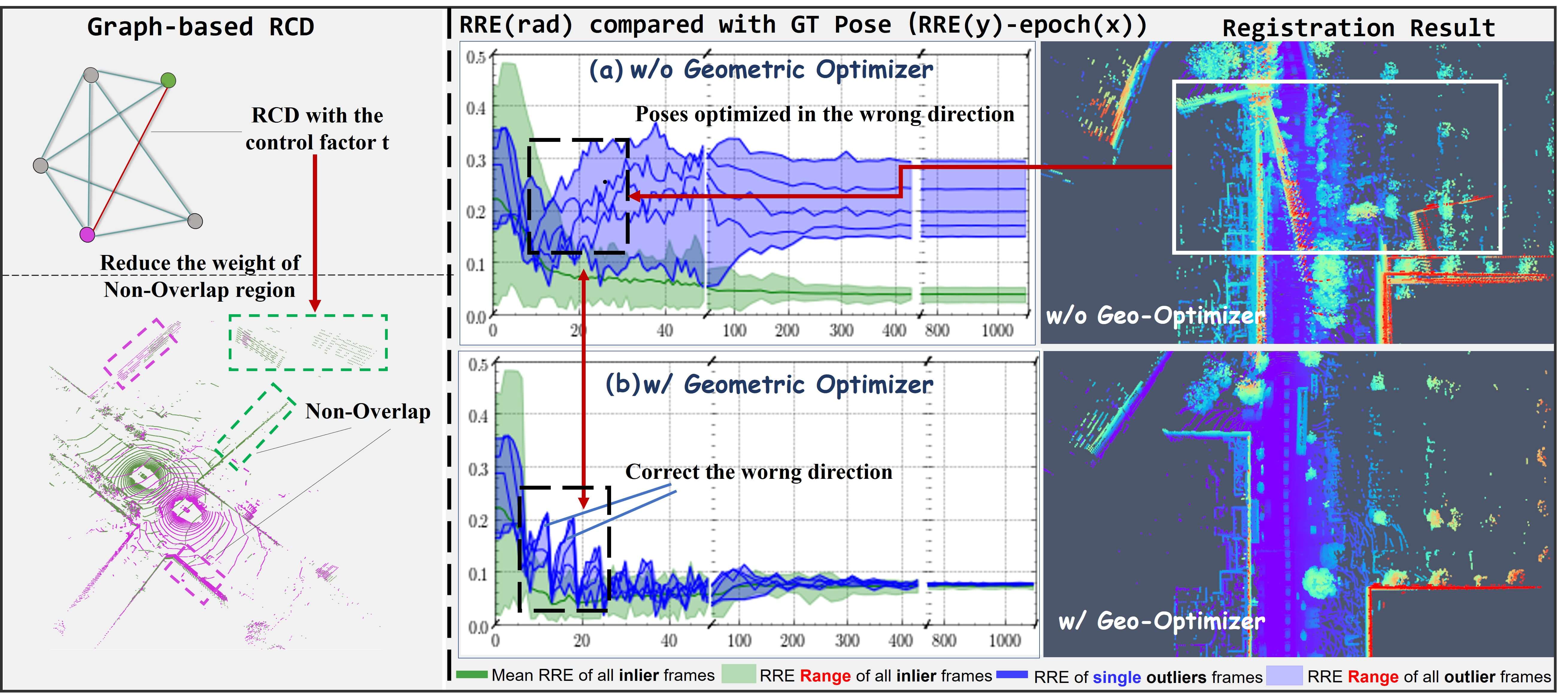}
  \captionsetup{aboveskip=1pt} 
  \caption{\textbf{Geo-optimizer and its impact on pose optimization}. Comparison of (a) and (b) shows Geo-optimizer prevents incorrect pose optimization of NeRF.}
  \label{fig:geooptim}
\end{figure*}  
Building upon the above, as shown in \cref{fig:geooptim}, we approximate the registration objective by optimizing the Graph-based Robust Chamfer distance (G-RCD). Specifically, 
we construct a graph $(\mathcal{S},\mathcal{E})$, where each vertex $\mathcal{S}$ represents a set of points and each edge corresponds to proposed RCD via \cref{eq: CD}. We connect each frame with its temporally adjacent $n$ frames to mitigate error accumulation in ICP~\cite{besl1992method}. Then RCD is calculated for all edges as \cref{eq: graph CD}, and $M$ denotes the number of frames in the sequence. Notably, in \cref{eq: CD}, G-RCD is computed using the global transform matrix, enabling direct gradient propagation of the Graph-based loss to the global transformation matrix of each frame.
\begin{equation}
\setlength\belowdisplayskip{0.05cm}
\label{eq: graph CD}
\begin{split}
    \mathcal{L}_{graph} =\frac{1}{(nM-\frac{n(n+1)}{2})} \sum_{(i,j) \in \mathcal{E}} \mathcal{L}_{(i,j)}, \
\end{split}
\end{equation}
\textbf{Discussion}. 
As illustrated in \cref{fig:geooptim}(b), insufficient geometric guidance leads to certain frame poses being optimized in the wrong direction. Geometric optimizer can address this issue by preventing pose updates strictly following NeRF and correcting wrong optimization directions that do not conform to global geometric consistency. This method involves externally modifying pose parameters and providing effective geometric guidance early in the ill-conditioned optimization process. Consequently, few iterations of graph-based RCD computation suffice to offer ample guidance for NeRF.



\subsection{Selective-Reweighting Strategy for Outlier Filtering}
\label{3.4}
\begin{figure*}[t]
\vspace{-.2cm}
\centering
\includegraphics[width=1\textwidth]{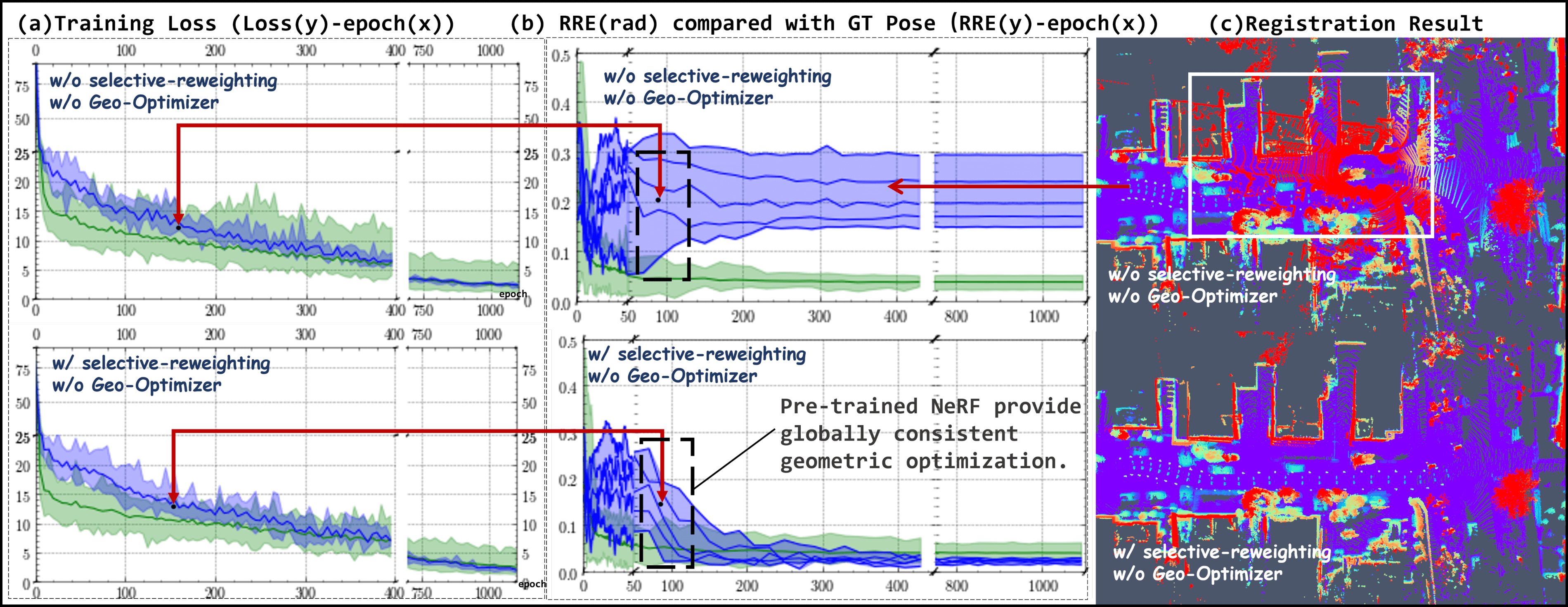}
  \captionsetup{aboveskip=1pt, belowskip=1pt} 
  \caption{\textbf{Impact of selective-reweighting training strategy on pose optimization.} (a) Frames with outlier poses exhibit significantly higher losses. With selective-reweighting, outlier frames maintain a relatively higher loss without overfitting. (b) After several training iterations, the pre-trained outlier-aware NeRF can provide globally consistent geometric optimization for outlier frames.}
  \label{fig:sec-rwt}
\vspace{-.3cm}
\end{figure*}  
In bundle-adjusting optimization, as shown in \cref{fig:sec-rwt}(a), we observed that frames with outlier poses present significantly higher rendering losses during the early stages of training. However, low frequency and sparsity of point clouds result in quick overfitting of individual frames including outliers (\textit{cf.} \cref{fig:sec-rwt}(a)(b)).
 This leads to minimal pose updates when the overall loss decreases, resulting in incorrect poses and inferior reconstruction.
Inspired by the capabilities of NeRF in pose inference~\cite{yen2021inerf}, we decrease the learning rate ($\mathtt{lr}$) of neural fields for the top k frames with the highest rendering losses as \cref{eq: reweight}, while keeping $\mathtt{lr}$ of poses unchanged. The strategy facilitates gradient propagation towards outlier poses, while the gradient flow to the radiance fields is concurrently diminished. 
Consequently, it's analogous to leveraging a pre-trained NeRF for outlier pose correction and lessens the adverse effects caused by outliers during the optimization process.
\begin{equation}
\label{eq: reweight}
    \mathrm{lr}_{outliers} = (w_0+l(1-w_0))\mathrm{lr}_{inliers} \quad
    (w_0>0)
\end{equation}
Where $l\in [0,1]$ denotes training progress. Akin to leaky ReLU~\cite{xu2020reluplex}, we set the reweighting factor $w_0$ to a relatively small value. $w_0$ increases as the process progresses, which ensures the network's ongoing learning from these frames and avoids stagnation. 
\subsection{Improving Geometry Constraints for NeRF}
\label{3.5}
Point clouds encapsulate rich geometric features. However, solely supervising NeRF training via range images pixel-wise fails to fully exploit their potential, \textit{e.g.}, normal information. Furthermore, the Chamfer distance can directly supervise the synthesized point clouds from a 3D perspective. Therefore, in addition to supervising via 2D range map, we propose directly constructing a three-dimensional geometric loss function between the generated point cloud and the ground truth point cloud. Unlike our Geo-optimizer, \cref{eq: CD2} imposes constraints between synthetic point clouds $\hat{P}$ and ground truth point clouds ${P}$:
\begin{small}
\begin{equation}
\setlength\belowdisplayskip{0.05cm}
\label{eq: CD2}
\begin{split}
    \mathcal{L}_{CD}=\frac{1}{N_{\hat{P}}} \sum_{\hat{p}_{i} \in \hat{P}} \min _{p_{i} \in P}\left\|\hat{p}_{i}-p_{i}\right\|^{2}_{2}
    +\frac{1}{N_P} \sum_{p_{i} \in P} \min _{\hat{p}_{i} \in \hat{P}}\left\|p_{i}-\hat{p}_{i}\right\|^{2}_{2}
\end{split}
\end{equation}
\end{small}
Based on the point correspondences established between $\hat{P}$ and $P$ as derived in equation (9), the constraint of normal can be formulated as minimizing:
\begin{small}
\begin{equation}
\setlength\belowdisplayskip{0.05cm}
\label{eq: normal}
\begin{split}
    \mathcal{L}_{normal}=
    \frac{1}{N_{\hat{P}}} \sum_{\hat{p}_{i} \in \hat{P}} \min _{p_{i} \in P}\left\|\mathcal{N}(\hat{p}_{i})-\mathcal{N}(p_{i})\right\|_{1}
    +\frac{1}{N_P} \sum_{p_{i} \in P} \min _{\hat{p}_{i} \in \hat{P}}\left\|\mathcal{N}(p_{i})-\mathcal{N}(\hat{p}_{i})\right\|_{1}
\end{split}
\end{equation}
\end{small}
Moreover, we also employ 2D loss function to supervise NeRF as \cref{nerf loss}.
\begin{small}
\begin{equation} \label{nerf loss}
  \mathcal{L}_{r}(\mathbf{r})
  =
  \sum_{\mathbf{r} \in \mathbf{R}} \lambda_d \begin{Vmatrix} \hat{D}(\mathbf{r}) - D(\mathbf{r}) \end{Vmatrix}_1 +
  \lambda_i \begin{Vmatrix} \hat{I}(\mathbf{r}) - I(\mathbf{r}) \end{Vmatrix}_2^2 + 
  \lambda_p \begin{Vmatrix} \hat{P}(\mathbf{r}) - P(\mathbf{r}) \end{Vmatrix}_2^2
\end{equation} 
\end{small}
Consequently, the loss for Neural LiDAR fields is weighted combination of the depth, intensity, ray-drop loss and 3D geometry constraints, which can be formalized as $\mathcal{L} = \mathcal{L}_{r} +\lambda_n \mathcal{L}_{normal} +\lambda_{c} \mathcal{L}_{CD}$.


\section{Experiment}
\label{sec:experiment}
\subsection{Experimental Setup}
\label{sec:setup}
\textbf{Datasets and Experimental Settings.}
We conducted experiments on two public autonomous driving datasets: NuScenes~\cite{caesar2020nuscenes} and KITTI-360~\cite{liao2022kitti} dataset, each with five representative LiDAR point cloud sequences.
We selected 36 consecutive frames at 2Hz from keyframes as a single scene for NuScenes, holding out 4 samples at 9-frame intervals for NVS evaluation. KITTI-360 has an acquisition frequency of 10Hz. We used 24 consecutive frames sampled every 5th frame to match scene sizes of Nuscenes, holding out 3 samples at 8-frame intervals for evaluation. We perturbed LiDAR poses with additive noise corresponding to a standard deviation of $20\deg$ in rotation and $3m$ in translation.

\textbf{Metrics.}
We evaluate our method from two perspectives: pose estimation and novel view synthesis. For pose evaluation, we use standard odometry metrics, including Absolute Trajectory Error ($\mathrm{ATE}$) and Relative Pose Error ($\mathrm{RPE_r}$ in rotation and $\mathrm{RPE_t}$ in translation). Following LiDAR4D~\cite{zheng2024lidar4d} for NVS evaluation, we employ CD to assess the 3D geometric error and the F-score with 5cm error threshold. Additionally, we use RMSE and MedAE to compute depth and intensity errors in projected range images, along with LPIPS~\cite{zhang2018unreasonable}, SSIM~\cite{wang2004image}, and PSNR to measure overall variance.

\textbf{Implementation Details.} 
The entire point cloud scene is scaled within the unit cube space. The optimization of GeoNLF is implemented on Pytorch~\cite{paszke2019pytorch} with Adam~\cite{kingma2014adam} optimizer. All the sequences are trained for 60K iterations. Our Geometry optimizer's $\mathtt{lr}$ for translation and rotation is the same as the $\mathtt{lr}$ for pose in NeRF with synchronized decay. We use the coarse-to-fine strategy\cite{lin2021barf,heo2023robust}, which starts from training progress 0.1 to 0.8. The reweight coefficient for the top-5 frames linearly increases from 0.15 to 1 during training. After every $m_1$ epoch of bundle adjusting global optimization, we proceed with $m_2$ epoch of pure geometric optimization, where $m_2/m_1$ decrease from 10 to 1. Please refer to \cref{apd:2} for more implementation details.
\subsection{Comparison in LiDAR NVS}
\label{sec:nvs}
\definecolor{best_result}{rgb}{0.96, 0.57, 0.58}
\definecolor{second_result}{rgb}{0.98, 0.78, 0.57}
\definecolor{third_result}{rgb}{1.0, 1.0, 0.56}
\begin{table*}[t]
\centering
\resizebox{\textwidth}{!}{
\renewcommand{\arraystretch}{1.35}
\begin{tabular}{ccccccccccccccc}
\hline \multirow[c]{2}{*}{ Method } & \multirow[c]{2}{*}{ Dataset } & \multicolumn{2}{c}{ Point Cloud } & \multicolumn{5}{c}{ Depth } & \multicolumn{5}{c}{ Intensity } \\
&    & CD$\downarrow$ & F-score$\uparrow$ & RMSE$\downarrow$ & MedAE$\downarrow$ & LPIPS$\downarrow$ & SSIM$\uparrow$ & PSNR$\uparrow$ & RMSE$\downarrow$ & MedAE$\downarrow$ & LPIPS$\downarrow$ & SSIM$\uparrow$ & PSNR$\uparrow$ \\
\hline
BARF-LN~\cite{lin2021barf,tao2023lidarnerf} & \multirow[c]{4}{*}{ Nuscenes }
& 1.2695 & 0.7589 & 8.2414 & 0.1123 & 0.1432 & \cellcolor{second_result}0.6856 & \cellcolor{second_result}20.89 & 0.392 & 0.0144 & 0.1023 & 0.6119 & 26.2330 \\
HASH-LN~\cite{heo2023robust,tao2023lidarnerf} &   
&\cellcolor{second_result}0.9691 & \cellcolor{second_result}0.8011 & \cellcolor{second_result}7.8353 & \cellcolor{second_result}0.0441 & \cellcolor{second_result}0.1190 & 0.6543 & 20.6244 & \cellcolor{second_result}0.0459 & \cellcolor{second_result}0.0135 & \cellcolor{second_result}0.0954 & \cellcolor{second_result}0.6279 & \cellcolor{second_result}26.8870 \\

GeoTrans~\cite{qin2022geometric,tao2023lidarnerf} &   
& 4.1587 & 0.2993 & 10.7899 & 2.1529 & 0.1445 & 0.3671 & 17.5885 & 0.0679 & 0.0256 & 0.1149 & 0.3476 & 23.6211\\

\textbf{GeoNLF~(Ours)} &   
& \cellcolor{best_result}\textbf{0.2408} & \cellcolor{best_result}\textbf{0.8647} & \cellcolor{best_result}\textbf{5.8208} & \cellcolor{best_result}\textbf{0.0281} & \cellcolor{best_result}\textbf{0.0727} & \cellcolor{best_result}\textbf{0.7746} & \cellcolor{best_result}\textbf{22.9472} & \cellcolor{best_result}\textbf{0.0378} & \cellcolor{best_result}\textbf{0.0100} & \cellcolor{best_result}\textbf{0.0774} & \cellcolor{best_result}\textbf{0.7368} & \cellcolor{best_result}\textbf{28.6078} \\
\hline
BAR-LN~\cite{lin2021barf,tao2023lidarnerf} & \multirow[c]{4}{*}{ KITTI-360 } 
& 3.1001 & 0.6156 & 7.5767 & 2.0583 & 0.5779 & 0.2834 & 22.5759 & \cellcolor{second_result}0.2121 & 0.1575 & 0.7121 & 0.1468 & 11.9778 \\

HASH-LN~\cite{heo2023robust,tao2023lidarnerf} &   
& 2.6913 & 0.6082 & 6.3005 & 2.1686 & 0.5176 & 0.3752 & 22.6196 & 0.2404 & \cellcolor{best_result}\textbf{0.1502} & 0.6508 & 0.1602 & 12.9286 \\

GeoTrans ~\cite{qin2022geometric,tao2023lidarnerf} &   
& \cellcolor{second_result}0.5753 & \cellcolor{second_result}0.8116 & \cellcolor{second_result}4.4291 & \cellcolor{second_result}0.2023 & \cellcolor{best_result}\textbf{0.3896} & \cellcolor{second_result}0.5330 & \cellcolor{best_result}\textbf{25.6137} & 0.2709 & 0.1589 & \cellcolor{second_result}0.5578 & \cellcolor{second_result}0.2578 & \cellcolor{second_result}12.9707\\

\textbf{GeoNLF~(Ours)} & 
& \cellcolor{best_result}\textbf{0.2363} & \cellcolor{best_result}\textbf{0.9178} & \cellcolor{best_result}\textbf{4.0293} & \cellcolor{best_result}\textbf{0.1009} & \cellcolor{second_result}0.3900 & \cellcolor{best_result}\textbf{0.6272} & \cellcolor{second_result}25.2758 & \cellcolor{best_result}\textbf{0.1495} & \cellcolor{second_result}0.1525 & \cellcolor{best_result}\textbf{0.5379} & \cellcolor{best_result}\textbf{0.3165} & \cellcolor{best_result}\textbf{16.5813} \\
\hline 

\end{tabular}}
\caption{\textbf{NVS Quantitative Comparison on Nuscenes and KITTI-360.} We compare our method to different types of approaches and color the top results as \colorbox{best_result}{best} and \colorbox{second_result}{second best}. All results are averaged over the 5 sequences.}
\label{exp:nvs_kn}
\vspace{-.2cm}
\end{table*}

\begin{figure*}[t]
\vspace{-.2cm}
\centering
\includegraphics[width=1\textwidth]{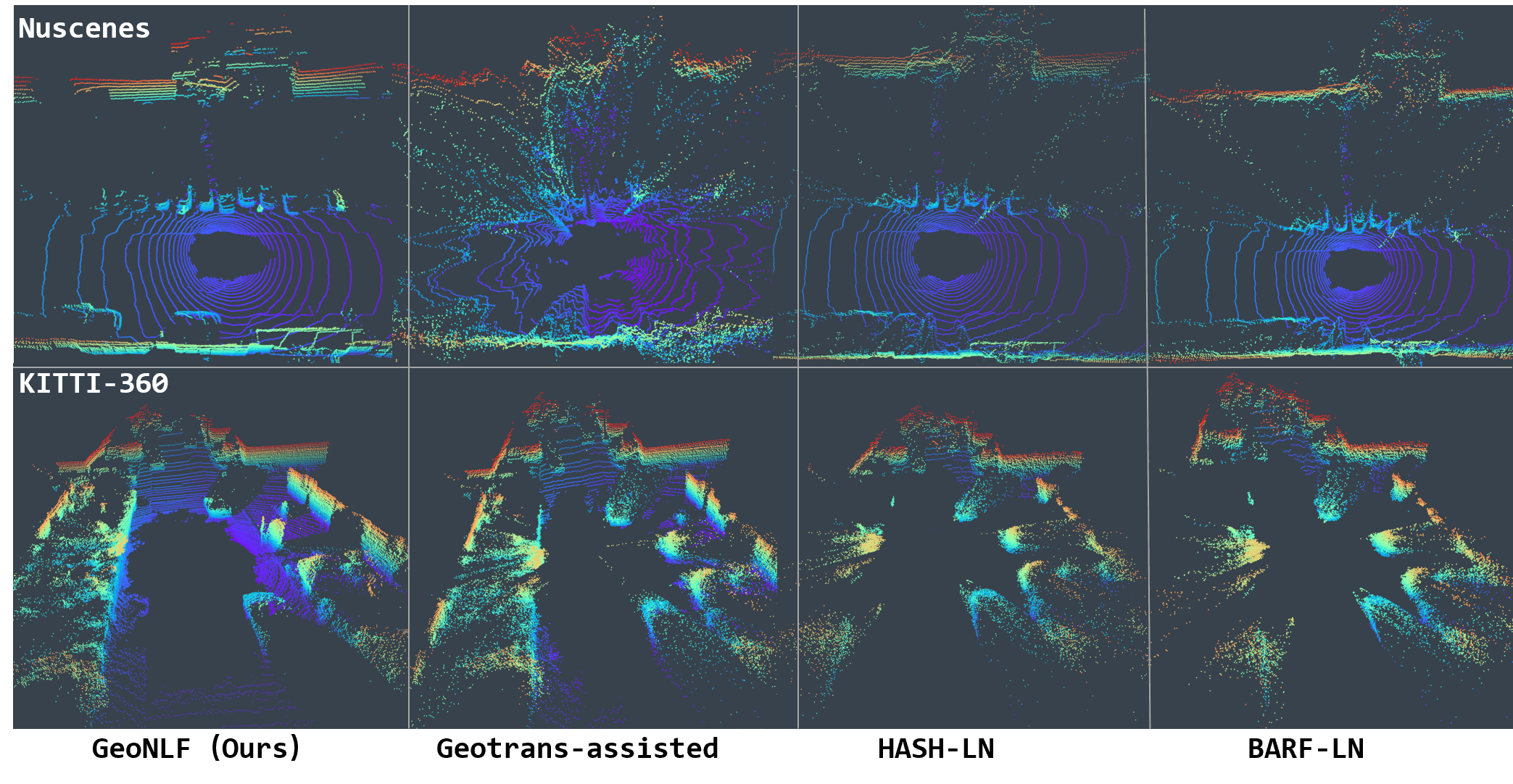}
  \caption{\textbf{Qualitative comparison of NVS.} We compared GeoNLF with other pose-free methods and GeoTrans-assisted NeRF. Especially, GeoTrans fails on Nuscenes due to the inaccurate poses. }
  \label{fig:nvs_kn}
\vspace{-.3cm}
\end{figure*}

We compare our model with BARF~\cite{lin2021barf} and HASH~\cite{heo2023robust}, both of which use LiDAR-NeRF\cite{tao2023lidarnerf} as backbone. For PCR-assisted NeRF, we opt to initially estimate pose utilizing pose derived from GeoTrans~\cite{qin2022geometric}, which is the most robust and accurate algorithm among other PCR methods in our experiments.
And subsequently we leverage LiDAR-NeRF~\cite{tao2023lidarnerf} for reconstruction. For all Pose-free methods, we follow NeRFmm\cite{wang2021nerf} to obtain the pose of test views for rendering.
The quantitative and qualitative results are in \cref{exp:nvs_kn} and \cref{fig:nvs_kn}.
Our method achieves high-precision registration and high-quality reconstruction across all sequences. However, baseline methods fail completely on certain sequences due to their lack of robustness. Please refer to \cref{apd:2} for details. Ultimately, our method excels in the reconstruction of depth and intensity, as evidenced by $7.9\%$ increase in F-score on Nuscenes and $13.1\%$ on KITTI-360 compared to the second best result.
\subsection{Comparison in Pose Estimation}
We conduct comprehensive comparisons of GeoNLF with pairwise baselines, including traditional method ICP~\cite{besl1992method}, learning-based GeoTrans~\cite{qin2022geometric} and outdoor-specific HRegNet~\cite{lu2021hregnet}, as well as multi-view baselines MICP~\cite{choi2015robust} and learning-based SGHR~\cite{wang2023robust}. For pairwise methods, we perform registration between adjacent frames in an Odometry-like way. For SGHR, we utilize FCGF~\cite{choy2019fully} descriptors followed by RANSAC~\cite{fischler1981random} for pairwise registration. The estimated trajectory is aligned with the ground truth using $\mathrm{Sim(3)}$ with known scale. Refer to \cref{apd:3} for details.

Our method dramatically outperforms both the registration and pose-free NeRF baselines. Quantitative results are illustrated in \cref{exp:reg_and_ab}. As depicted in \cref{fig:reg_res}, most registration methods fail to achieve globally accurate poses and \textbf{completely fail in some scenarios}, leading to massive errors in final average results. Significant generalization issues arise for learning-based registration methods due to potential disparities between testing scenarios and training data, including differences in initial pose distributions. This challenge is particularly pronounced in HRegNet~\cite{lu2021hregnet}. While the transformer model GeoTrans~\cite{qin2022geometric} with its higher capacity offers some alleviation to the issue, it remains not fully resolved. We provide more specific results and analysis in \cref{apd:3} for further clarification.
\subsection{Ablation Study}
\label{sec:ablation}
\begin{figure*}[t]
\vspace{-.1cm}
\centering
\includegraphics[width=1\textwidth]{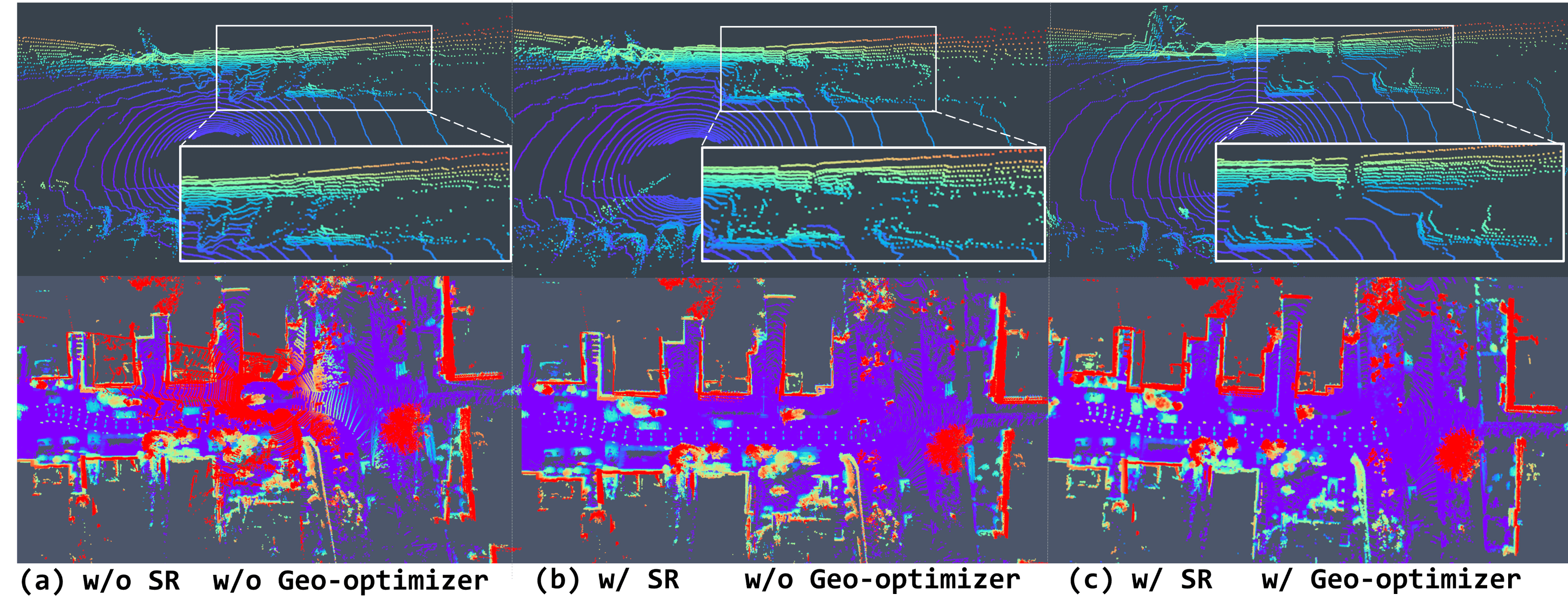}
  \caption{\textbf{Qualitative results of ablation study.} We present the NVS and Registration results in the first and second rows. Outlier frames emerged w/o SR or w/o G-optim.}
  \label{fig:abla}
\vspace{-.3cm}
\end{figure*}

In this Section, we analyze the effectiveness of each component of GeoNLF. The results of ablation studies are shown in \cref{exp:reg_and_ab}.
\textbf{(1) Geo-optimizer.} When training GeoNLF w/o geo-optimizer (w/o G-optim), pose optimization may initially converge towards incorrect directions. Excluding geo-optimizer in GeoNLF results in decreased pose accuracy and reconstruction quality. \textbf{(2) Control factor of graph-based RCD.} Although geo-optimizer is crucial in the early stages of optimization, we find that using the original CD limits the accuracy of pose estimation. Removing the control factor (w/o RCD) leads to decreased pose estimation accuracy due to the presence of non-overlapping regions. \textbf{(3) Selective-reweighting (SR) strategy.} As presented in \cref{fig:abla,fig:sec-rwt} and \cref{exp:reg_and_ab}, outlier frames cause GeoNLF w/o SR strategy to overlook multi-view consistency, adversely affecting reconstruction quality. \textbf{(4) Geometric constraints.} Removing the 3D constraints (w/o $L_{3d}$) results in a decline in CD due to the photometric loss's inability to adequately capture geometric information.
\vspace{-0.2cm}
\definecolor{best_result}{rgb}{0.96, 0.57, 0.58}
\definecolor{second_result}{rgb}{0.98, 0.78, 0.57}
\definecolor{third_result}{rgb}{1.0, 1.0, 0.56}
\begin{table*}[t]
\vspace{-1.1cm}
\begin{minipage}{0.51\textwidth}
\centering
\resizebox{\textwidth}{!}{
\renewcommand{\arraystretch}{1.5}
\begin{tabular}{ccccccccccccccc}
\hline \multirow[c]{2}{*}{ Method }& \multicolumn{3}{c}{ NuScenes } & \multicolumn{3}{c}{ KITTI-360 } \\
& $\mathrm{RPE_t}$(cm)$\downarrow$ & $\mathrm{RPE_r}$(deg)$\downarrow$ & $\mathrm{ATE}$(m)$\downarrow$ & $\mathrm{RPE_t}$(cm)$\downarrow$ & $\mathrm{RPE_r}$(deg) $\downarrow$ & $\mathrm{ATE}$(m) $\downarrow$ \\

\hline ICP~\cite{besl1992method}  & \cellcolor{second_result}15.410 & 0.647 & 1.131 & 30.383 & 1.019 & 1.894 \\

MICP~\cite{tao2023lidarnerf} &  38.84 & 1.101 & 2.519 & 35.584 & 1.419 & 1.483 \\

HRegNet~\cite{lu2021hregnet} &  120.913 & 2.173 & 7.815 & 290.16 & 9.083 & 7.423  \\

SGHR~\cite{wang2023robust} &  100.98 & 0.699 & 9.557 & 95.576 & 0.906 & 2.539 \\

GeoTrans~\cite{qin2022geometric} &  16.097 &  \cellcolor{second_result}0.363 &  \cellcolor{second_result}0.892 &  \cellcolor{second_result}6.081 &  \cellcolor{second_result}0.213 &  \cellcolor{second_result}0.246  \\

BARF-LN~\cite{tao2023lidarnerf,lin2021barf}  & 210.331 & 0.819 & 5.244 & 199.74 & 2.203 & 2.763 \\

HASH-LN~\cite{tao2023lidarnerf,heo2023robust} &  180.282 & 0.832 & 4.151 & 196.791 & 2.171 & 2.666 \\

\textbf{GeoNLF (Ours)}  & \cellcolor{best_result}\textbf{7.058} & \cellcolor{best_result}\textbf{0.103} & \cellcolor{best_result}\textbf{0.228} & \cellcolor{best_result}\textbf{5.449} & \cellcolor{best_result}\textbf{0.205} & \cellcolor{best_result}\textbf{0.170}\\
\hline
\end{tabular}}
\caption{\textbf{Pose estimation accuracy comparison.}}  
\label{exp:reg}
\end{minipage}
\hfill
\begin{minipage}{0.48\textwidth}
\centering
\resizebox{\textwidth}{!}{
\renewcommand{\arraystretch}{2}
\begin{tabular}{ccccccccccccccc}
\hline 
\multirow[c]{2}{*}{ Method }&\multicolumn{1}{c}{Point Cloud} & \multicolumn{1}{c}{Depth} & \multicolumn{1}{c}{Intensity} &  \multicolumn{3}{c}{Pose}\\
& CD$\downarrow$ & $\mathrm{PSNR}\uparrow$ & $\mathrm{PSNR}\uparrow$ & $\mathrm{RPE_t}$(cm)$\downarrow$ &$\mathrm{RPE_r}$(deg)$\downarrow$ &ATE(m)$\downarrow$ \\
\hline 
w/o G-optim   & 0.6180 & 21.3211 & 25.8551    &  54.72  & 0.283  & 1.328 \\
w/o RCD       & 0.2711  & 21.1323 & 26.7232    &  8.476 & 0.163  & 0.332 \\
w/o SR        & 0.2654  & 21.1096 & 26.5269    &  8.124 & 0.156  & 0.264 \\
w/o $L_{3d}$  & 0.2877    & 21.7128 & 28.5210     &  7.273 & 0.124  & 0.234 \\
\textbf{GeoNLF} &\textbf{0.2363}&\textbf{22.9472}&\textbf{28.6078}& \textbf{7.058}&\textbf{0.103}& \textbf{0.228} \\
\hline
\end{tabular}}
\caption{\textbf{Ablation study on Nuscenes.}}  
\label{exp:reg_and_ab}
\end{minipage}
\vspace{-.3cm}
\end{table*}

\vspace{-0.3cm}
\subsection{Limination}
Despite the fact that GeoNLF has exhibited exceptional performance in PCR and LiDAR-NVS on challenging scenes, it is not designed for dynamic scenes, which is non-negligible in autonomous driving scenarios. Additionally, GeoNLF targets point clouds within a sequence, relying on the temporal prior of the point clouds.

\section{Conclusion}
\label{sec:conclusion}
We introduce GeoNLF for multi-view registration and novel view synthesis from a sequence of sparsely sampled point clouds. We demonstrate the challenges encountered by previous pairwise and multi-view registration methods, as well as the difficulties faced by previous pose-free methods. Through the utilization of our Geo-Optimizer, Graph-based Robust CD, selective-reweighting strategy and geometric constraints from 3D perspective, our outlier-aware and geometry-aware GeoNLF demonstrate the promising performance in both multi-view registration and NVS tasks.

\bibliographystyle{plainnat}
\bibliography{ref}

\begin{thebibliography}{71}
\providecommand{\natexlab}[1]{#1}
\providecommand{\url}[1]{\texttt{#1}}
\expandafter\ifx\csname urlstyle\endcsname\relax
  \providecommand{\doi}[1]{doi: #1}\else
  \providecommand{\doi}{doi: \begingroup \urlstyle{rm}\Url}\fi

\bibitem[Aoki et~al.(2019)Aoki, Goforth, Srivatsan, and Lucey]{aoki2019pointnetlk}
Yasuhiro Aoki, Hunter Goforth, Rangaprasad~Arun Srivatsan, and Simon Lucey.
\newblock Pointnetlk: Robust \& efficient point cloud registration using pointnet.
\newblock In \emph{Proceedings of the IEEE/CVF conference on computer vision and pattern recognition}, pages 7163--7172, 2019.

\bibitem[Arie-Nachimson et~al.(2012)Arie-Nachimson, Kovalsky, Kemelmacher-Shlizerman, Singer, and Basri]{arie2012global}
Mica Arie-Nachimson, Shahar~Z Kovalsky, Ira Kemelmacher-Shlizerman, Amit Singer, and Ronen Basri.
\newblock Global motion estimation from point matches.
\newblock In \emph{2012 Second international conference on 3D imaging, modeling, processing, visualization \& transmission}, pages 81--88. IEEE, 2012.

\bibitem[Arrigoni et~al.(2016)Arrigoni, Rossi, and Fusiello]{arrigoni2016spectral}
Federica Arrigoni, Beatrice Rossi, and Andrea Fusiello.
\newblock Spectral synchronization of multiple views in se (3).
\newblock \emph{SIAM Journal on Imaging Sciences}, 9\penalty0 (4):\penalty0 1963--1990, 2016.

\bibitem[Barron et~al.(2021)Barron, Mildenhall, Tancik, Hedman, Martin-Brualla, and Srinivasan]{barron2021mip}
Jonathan~T Barron, Ben Mildenhall, Matthew Tancik, Peter Hedman, Ricardo Martin-Brualla, and Pratul~P Srinivasan.
\newblock Mip-nerf: A multiscale representation for anti-aliasing neural radiance fields.
\newblock In \emph{Proceedings of the IEEE/CVF International Conference on Computer Vision}, pages 5855--5864, 2021.

\bibitem[Besl and McKay(1992)]{besl1992method}
Paul~J Besl and Neil~D McKay.
\newblock Method for registration of 3-d shapes.
\newblock In \emph{Sensor fusion IV: control paradigms and data structures}, volume 1611, pages 586--606. Spie, 1992.

\bibitem[Bian et~al.(2023)Bian, Wang, Li, Bian, and Prisacariu]{bian2023nope}
Wenjing Bian, Zirui Wang, Kejie Li, Jia-Wang Bian, and Victor~Adrian Prisacariu.
\newblock Nope-nerf: Optimising neural radiance field with no pose prior.
\newblock In \emph{Proceedings of the IEEE/CVF Conference on Computer Vision and Pattern Recognition}, pages 4160--4169, 2023.

\bibitem[Birdal and Ilic(2017)]{birdal2017cad}
Tolga Birdal and Slobodan Ilic.
\newblock Cad priors for accurate and flexible instance reconstruction.
\newblock In \emph{Proceedings of the IEEE international conference on computer vision}, pages 133--142, 2017.

\bibitem[Birdal and Simsekli(2019)]{birdal2019probabilistic}
Tolga Birdal and Umut Simsekli.
\newblock Probabilistic permutation synchronization using the riemannian structure of the birkhoff polytope.
\newblock In \emph{Proceedings of the IEEE/CVF Conference on Computer Vision and Pattern Recognition}, pages 11105--11116, 2019.

\bibitem[Caesar et~al.(2020)Caesar, Bankiti, Lang, Vora, Liong, Xu, Krishnan, Pan, Baldan, and Beijbom]{caesar2020nuscenes}
Holger Caesar, Varun Bankiti, Alex~H Lang, Sourabh Vora, Venice~Erin Liong, Qiang Xu, Anush Krishnan, Yu~Pan, Giancarlo Baldan, and Oscar Beijbom.
\newblock nuscenes: A multimodal dataset for autonomous driving.
\newblock In \emph{Proceedings of the IEEE/CVF conference on computer vision and pattern recognition}, pages 11621--11631, 2020.

\bibitem[Chan et~al.(2022)Chan, Lin, Chan, Nagano, Pan, De~Mello, Gallo, Guibas, Tremblay, Khamis, et~al.]{chan2022efficient}
Eric~R Chan, Connor~Z Lin, Matthew~A Chan, Koki Nagano, Boxiao Pan, Shalini De~Mello, Orazio Gallo, Leonidas~J Guibas, Jonathan Tremblay, Sameh Khamis, et~al.
\newblock Efficient geometry-aware 3d generative adversarial networks.
\newblock In \emph{Proceedings of the IEEE/CVF conference on computer vision and pattern recognition}, pages 16123--16133, 2022.

\bibitem[Chen et~al.(2022)Chen, Xu, Geiger, Yu, and Su]{chen2022tensorf}
Anpei Chen, Zexiang Xu, Andreas Geiger, Jingyi Yu, and Hao Su.
\newblock Tensorf: Tensorial radiance fields.
\newblock In \emph{European Conference on Computer Vision}, pages 333--350. Springer, 2022.

\bibitem[Chen and Lee(2023)]{chen2023dbarf}
Yu~Chen and Gim~Hee Lee.
\newblock Dbarf: Deep bundle-adjusting generalizable neural radiance fields.
\newblock In \emph{Proceedings of the IEEE/CVF Conference on Computer Vision and Pattern Recognition}, pages 24--34, 2023.

\bibitem[Choi et~al.(2015)Choi, Zhou, and Koltun]{choi2015robust}
Sungjoon Choi, Qian-Yi Zhou, and Vladlen Koltun.
\newblock Robust reconstruction of indoor scenes.
\newblock In \emph{Proceedings of the IEEE conference on computer vision and pattern recognition}, pages 5556--5565, 2015.

\bibitem[Choy et~al.(2019)Choy, Park, and Koltun]{choy2019fully}
Christopher Choy, Jaesik Park, and Vladlen Koltun.
\newblock Fully convolutional geometric features.
\newblock In \emph{Proceedings of the IEEE/CVF international conference on computer vision}, pages 8958--8966, 2019.

\bibitem[Deng et~al.(2023)Deng, Wu, Chen, Xia, Sun, Liu, Yu, and Pei]{deng2023nerf}
Junyuan Deng, Qi~Wu, Xieyuanli Chen, Songpengcheng Xia, Zhen Sun, Guoqing Liu, Wenxian Yu, and Ling Pei.
\newblock Nerf-loam: Neural implicit representation for large-scale incremental lidar odometry and mapping.
\newblock In \emph{Proceedings of the IEEE/CVF International Conference on Computer Vision}, pages 8218--8227, 2023.

\bibitem[Deng et~al.(2022)Deng, Liu, Zhu, and Ramanan]{deng2022depth}
Kangle Deng, Andrew Liu, Jun-Yan Zhu, and Deva Ramanan.
\newblock Depth-supervised nerf: Fewer views and faster training for free.
\newblock In \emph{Proceedings of the IEEE/CVF Conference on Computer Vision and Pattern Recognition}, pages 12882--12891, 2022.

\bibitem[Dosovitskiy et~al.(2017)Dosovitskiy, Ros, Codevilla, Lopez, and Koltun]{dosovitskiy2017carla}
Alexey Dosovitskiy, German Ros, Felipe Codevilla, Antonio Lopez, and Vladlen Koltun.
\newblock Carla: An open urban driving simulator.
\newblock In \emph{Conference on robot learning}, pages 1--16. PMLR, 2017.

\bibitem[Fischler and Bolles(1981)]{fischler1981random}
Martin~A Fischler and Robert~C Bolles.
\newblock Random sample consensus: a paradigm for model fitting with applications to image analysis and automated cartography.
\newblock \emph{Communications of the ACM}, 24\penalty0 (6):\penalty0 381--395, 1981.

\bibitem[Geiger et~al.(2012)Geiger, Lenz, and Urtasun]{geiger2012we}
Andreas Geiger, Philip Lenz, and Raquel Urtasun.
\newblock Are we ready for autonomous driving? the kitti vision benchmark suite.
\newblock In \emph{2012 IEEE conference on computer vision and pattern recognition}, pages 3354--3361. IEEE, 2012.

\bibitem[Gojcic et~al.(2020)Gojcic, Zhou, Wegner, Guibas, and Birdal]{gojcic2020learning}
Zan Gojcic, Caifa Zhou, Jan~D Wegner, Leonidas~J Guibas, and Tolga Birdal.
\newblock Learning multiview 3d point cloud registration.
\newblock In \emph{Proceedings of the IEEE/CVF conference on computer vision and pattern recognition}, pages 1759--1769, 2020.

\bibitem[Guillard et~al.(2022)Guillard, Vemprala, Gupta, Miksik, Vineet, Fua, and Kapoor]{guillard2022learning}
Beno{\^\i}t Guillard, Sai Vemprala, Jayesh~K Gupta, Ondrej Miksik, Vibhav Vineet, Pascal Fua, and Ashish Kapoor.
\newblock Learning to simulate realistic lidars.
\newblock In \emph{2022 IEEE/RSJ International Conference on Intelligent Robots and Systems (IROS)}, pages 8173--8180. IEEE, 2022.

\bibitem[Heo et~al.(2023)Heo, Kim, Lee, Lee, Kim, Kim, and Kim]{heo2023robust}
Hwan Heo, Taekyung Kim, Jiyoung Lee, Jaewon Lee, Soohyun Kim, Hyunwoo~J Kim, and Jin-Hwa Kim.
\newblock Robust camera pose refinement for multi-resolution hash encoding.
\newblock In \emph{International Conference on Machine Learning}, pages 13000--13016. PMLR, 2023.

\bibitem[Hu et~al.(2023)Hu, Wang, Ma, Yang, Gao, Liu, and Ma]{hu2023tri}
Wenbo Hu, Yuling Wang, Lin Ma, Bangbang Yang, Lin Gao, Xiao Liu, and Yuewen Ma.
\newblock Tri-miprf: Tri-mip representation for efficient anti-aliasing neural radiance fields.
\newblock In \emph{Proceedings of the IEEE/CVF International Conference on Computer Vision}, pages 19774--19783, 2023.

\bibitem[Huang et~al.(2023)Huang, Gojcic, Wang, Williams, Kasten, Fidler, Schindler, and Litany]{huang2023nfl}
Shengyu Huang, Zan Gojcic, Zian Wang, Francis Williams, Yoni Kasten, Sanja Fidler, Konrad Schindler, and Or~Litany.
\newblock Neural lidar fields for novel view synthesis.
\newblock \emph{arXiv preprint arXiv:2305.01643}, 2023.

\bibitem[Huang et~al.(2020)Huang, Mei, and Zhang]{huang2020feature}
Xiaoshui Huang, Guofeng Mei, and Jian Zhang.
\newblock Feature-metric registration: A fast semi-supervised approach for robust point cloud registration without correspondences.
\newblock In \emph{Proceedings of the IEEE/CVF conference on computer vision and pattern recognition}, pages 11366--11374, 2020.

\bibitem[Jin et~al.(2024)Jin, Armeni, Pollefeys, and Barath]{jin2024multiway}
Shengze Jin, Iro Armeni, Marc Pollefeys, and Daniel Barath.
\newblock Multiway point cloud mosaicking with diffusion and global optimization.
\newblock \emph{arXiv preprint arXiv:2404.00429}, 2024.

\bibitem[Kingma and Ba(2014)]{kingma2014adam}
Diederik~P Kingma and Jimmy Ba.
\newblock Adam: A method for stochastic optimization.
\newblock \emph{arXiv preprint arXiv:1412.6980}, 2014.

\bibitem[Koenig and Howard(2004)]{koenig2004design}
Nathan Koenig and Andrew Howard.
\newblock Design and use paradigms for gazebo, an open-source multi-robot simulator.
\newblock In \emph{2004 IEEE/RSJ international conference on intelligent robots and systems (IROS)(IEEE Cat. No. 04CH37566)}, volume~3, pages 2149--2154. Ieee, 2004.

\bibitem[Li et~al.(2023)Li, Ren, and Liu]{li2023pcgen}
Chenqi Li, Yuan Ren, and Bingbing Liu.
\newblock Pcgen: Point cloud generator for lidar simulation.
\newblock In \emph{2023 IEEE International Conference on Robotics and Automation (ICRA)}, pages 11676--11682. IEEE, 2023.

\bibitem[Li et~al.(2020)Li, Zhang, Xu, Zhou, and Zhang]{li2020iterative}
Jiahao Li, Changhao Zhang, Ziyao Xu, Hangning Zhou, and Chi Zhang.
\newblock Iterative distance-aware similarity matrix convolution with mutual-supervised point elimination for efficient point cloud registration.
\newblock In \emph{Computer Vision--ECCV 2020: 16th European Conference, Glasgow, UK, August 23--28, 2020, Proceedings, Part XXIV 16}, pages 378--394. Springer, 2020.

\bibitem[Liao et~al.(2022)Liao, Xie, and Geiger]{liao2022kitti}
Yiyi Liao, Jun Xie, and Andreas Geiger.
\newblock Kitti-360: A novel dataset and benchmarks for urban scene understanding in 2d and 3d.
\newblock \emph{IEEE Transactions on Pattern Analysis and Machine Intelligence}, 45\penalty0 (3):\penalty0 3292--3310, 2022.

\bibitem[Lin et~al.(2021)Lin, Ma, Torralba, and Lucey]{lin2021barf}
Chen-Hsuan Lin, Wei-Chiu Ma, Antonio Torralba, and Simon Lucey.
\newblock Barf: Bundle-adjusting neural radiance fields.
\newblock In \emph{Proceedings of the IEEE/CVF International Conference on Computer Vision}, pages 5741--5751, 2021.

\bibitem[Lin et~al.(2023)Lin, M{\"u}ller, Tremblay, Wen, Tyree, Evans, Vela, and Birchfield]{lin2023parallel}
Yunzhi Lin, Thomas M{\"u}ller, Jonathan Tremblay, Bowen Wen, Stephen Tyree, Alex Evans, Patricio~A Vela, and Stan Birchfield.
\newblock Parallel inversion of neural radiance fields for robust pose estimation.
\newblock In \emph{2023 IEEE International Conference on Robotics and Automation (ICRA)}, pages 9377--9384. IEEE, 2023.

\bibitem[Liu et~al.(2023)Liu, Gao, Meuleman, Tseng, Saraf, Kim, Chuang, Kopf, and Huang]{liu2023robust}
Yu-Lun Liu, Chen Gao, Andreas Meuleman, Hung-Yu Tseng, Ayush Saraf, Changil Kim, Yung-Yu Chuang, Johannes Kopf, and Jia-Bin Huang.
\newblock Robust dynamic radiance fields.
\newblock In \emph{Proceedings of the IEEE/CVF Conference on Computer Vision and Pattern Recognition}, pages 13--23, 2023.

\bibitem[Lu et~al.(2021)Lu, Chen, Liu, Zhang, Qu, Liu, and Gu]{lu2021hregnet}
Fan Lu, Guang Chen, Yinlong Liu, Lijun Zhang, Sanqing Qu, Shu Liu, and Rongqi Gu.
\newblock Hregnet: A hierarchical network for large-scale outdoor lidar point cloud registration.
\newblock In \emph{Proceedings of the IEEE/CVF International Conference on Computer Vision}, pages 16014--16023, 2021.

\bibitem[Manivasagam et~al.(2020)Manivasagam, Wang, Wong, Zeng, Sazanovich, Tan, Yang, Ma, and Urtasun]{manivasagam2020lidarsim}
Sivabalan Manivasagam, Shenlong Wang, Kelvin Wong, Wenyuan Zeng, Mikita Sazanovich, Shuhan Tan, Bin Yang, Wei-Chiu Ma, and Raquel Urtasun.
\newblock Lidarsim: Realistic lidar simulation by leveraging the real world.
\newblock In \emph{Proceedings of the IEEE/CVF Conference on Computer Vision and Pattern Recognition}, pages 11167--11176, 2020.

\bibitem[Maset et~al.(2017)Maset, Arrigoni, and Fusiello]{maset2017practical}
Eleonora Maset, Federica Arrigoni, and Andrea Fusiello.
\newblock Practical and efficient multi-view matching.
\newblock In \emph{Proceedings of the IEEE International Conference on Computer Vision}, pages 4568--4576, 2017.

\bibitem[Mildenhall et~al.(2021)Mildenhall, Srinivasan, Tancik, Barron, Ramamoorthi, and Ng]{mildenhall2021nerf}
Ben Mildenhall, Pratul~P Srinivasan, Matthew Tancik, Jonathan~T Barron, Ravi Ramamoorthi, and Ren Ng.
\newblock Nerf: Representing scenes as neural radiance fields for view synthesis.
\newblock \emph{Communications of the ACM}, 65\penalty0 (1):\penalty0 99--106, 2021.

\bibitem[M{\"u}ller et~al.(2022)M{\"u}ller, Evans, Schied, and Keller]{muller2022instant}
Thomas M{\"u}ller, Alex Evans, Christoph Schied, and Alexander Keller.
\newblock Instant neural graphics primitives with a multiresolution hash encoding.
\newblock \emph{ACM transactions on graphics (TOG)}, 41\penalty0 (4):\penalty0 1--15, 2022.

\bibitem[Niemeyer et~al.(2022)Niemeyer, Barron, Mildenhall, Sajjadi, Geiger, and Radwan]{niemeyer2022regnerf}
Michael Niemeyer, Jonathan~T Barron, Ben Mildenhall, Mehdi~SM Sajjadi, Andreas Geiger, and Noha Radwan.
\newblock Regnerf: Regularizing neural radiance fields for view synthesis from sparse inputs.
\newblock In \emph{Proceedings of the IEEE/CVF Conference on Computer Vision and Pattern Recognition}, pages 5480--5490, 2022.

\bibitem[Oechsle et~al.(2021)Oechsle, Peng, and Geiger]{oechsle2021unisurf}
Michael Oechsle, Songyou Peng, and Andreas Geiger.
\newblock Unisurf: Unifying neural implicit surfaces and radiance fields for multi-view reconstruction.
\newblock In \emph{Proceedings of the IEEE/CVF International Conference on Computer Vision}, pages 5589--5599, 2021.

\bibitem[Park et~al.(2023)Park, Henzler, Mildenhall, Barron, and Martin-Brualla]{park2023camp}
Keunhong Park, Philipp Henzler, Ben Mildenhall, Jonathan~T Barron, and Ricardo Martin-Brualla.
\newblock Camp: Camera preconditioning for neural radiance fields.
\newblock \emph{ACM Transactions on Graphics (TOG)}, 42\penalty0 (6):\penalty0 1--11, 2023.

\bibitem[Paszke et~al.(2019)Paszke, Gross, Massa, Lerer, Bradbury, Chanan, Killeen, Lin, Gimelshein, Antiga, et~al.]{paszke2019pytorch}
Adam Paszke, Sam Gross, Francisco Massa, Adam Lerer, James Bradbury, Gregory Chanan, Trevor Killeen, Zeming Lin, Natalia Gimelshein, Luca Antiga, et~al.
\newblock Pytorch: An imperative style, high-performance deep learning library.
\newblock \emph{Advances in neural information processing systems}, 32, 2019.

\bibitem[Pomerleau et~al.(2013)Pomerleau, Colas, Siegwart, and Magnenat]{pomerleau2013comparing}
Fran{\c{c}}ois Pomerleau, Francis Colas, Roland Siegwart, and St{\'e}phane Magnenat.
\newblock Comparing icp variants on real-world data sets: Open-source library and experimental protocol.
\newblock \emph{Autonomous robots}, 34:\penalty0 133--148, 2013.

\bibitem[Qin et~al.(2022)Qin, Yu, Wang, Guo, Peng, and Xu]{qin2022geometric}
Zheng Qin, Hao Yu, Changjian Wang, Yulan Guo, Yuxing Peng, and Kai Xu.
\newblock Geometric transformer for fast and robust point cloud registration.
\newblock In \emph{Proceedings of the IEEE/CVF conference on computer vision and pattern recognition}, pages 11143--11152, 2022.

\bibitem[Ramalingam and Taguchi(2013)]{ramalingam2013theory}
Srikumar Ramalingam and Yuichi Taguchi.
\newblock A theory of minimal 3d point to 3d plane registration and its generalization.
\newblock \emph{International journal of computer vision}, 102:\penalty0 73--90, 2013.

\bibitem[Roessle et~al.(2022)Roessle, Barron, Mildenhall, Srinivasan, and Nie{\ss}ner]{roessle2022dense}
Barbara Roessle, Jonathan~T Barron, Ben Mildenhall, Pratul~P Srinivasan, and Matthias Nie{\ss}ner.
\newblock Dense depth priors for neural radiance fields from sparse input views.
\newblock In \emph{Proceedings of the IEEE/CVF Conference on Computer Vision and Pattern Recognition}, pages 12892--12901, 2022.

\bibitem[Rusinkiewicz and Levoy(2001)]{rusinkiewicz2001efficient}
Szymon Rusinkiewicz and Marc Levoy.
\newblock Efficient variants of the icp algorithm.
\newblock In \emph{Proceedings third international conference on 3-D digital imaging and modeling}, pages 145--152. IEEE, 2001.

\bibitem[Schonberger and Frahm(2016)]{schonberger2016structure}
Johannes~L Schonberger and Jan-Michael Frahm.
\newblock Structure-from-motion revisited.
\newblock In \emph{Proceedings of the IEEE conference on computer vision and pattern recognition}, pages 4104--4113, 2016.

\bibitem[Shah et~al.(2018)Shah, Dey, Lovett, and Kapoor]{shah2018airsim}
Shital Shah, Debadeepta Dey, Chris Lovett, and Ashish Kapoor.
\newblock Airsim: High-fidelity visual and physical simulation for autonomous vehicles.
\newblock In \emph{Field and Service Robotics: Results of the 11th International Conference}, pages 621--635. Springer, 2018.

\bibitem[Song et~al.(2023)Song, Wang, Liu, Fu, Miao, et~al.]{song2023sc}
Liang Song, Guangming Wang, Jiuming Liu, Zhenyang Fu, Yanzi Miao, et~al.
\newblock Sc-nerf: Self-correcting neural radiance field with sparse views.
\newblock \emph{arXiv preprint arXiv:2309.05028}, 2023.

\bibitem[Tao et~al.(2023)Tao, Gao, Wang, Chen, Hao, Liang, Salzmann, and Yu]{tao2023lidarnerf}
Tang Tao, Longfei Gao, Guangrun Wang, Peng Chen, Dayang Hao, Xiaodan Liang, Mathieu Salzmann, and Kaicheng Yu.
\newblock Lidar-nerf: Novel lidar view synthesis via neural radiance fields.
\newblock \emph{arXiv preprint arXiv:2304.10406}, 2023.

\bibitem[Tejus et~al.(2023)Tejus, Zara, Rota, Fusiello, Ricci, and Arrigoni]{tejus2023rotation}
GK~Tejus, Giacomo Zara, Paolo Rota, Andrea Fusiello, Elisa Ricci, and Federica Arrigoni.
\newblock Rotation synchronization via deep matrix factorization.
\newblock In \emph{2023 IEEE International Conference on Robotics and Automation (ICRA)}, pages 2113--2119. IEEE, 2023.

\bibitem[Truong et~al.(2023)Truong, Rakotosaona, Manhardt, and Tombari]{truong2023sparf}
Prune Truong, Marie-Julie Rakotosaona, Fabian Manhardt, and Federico Tombari.
\newblock Sparf: Neural radiance fields from sparse and noisy poses.
\newblock In \emph{Proceedings of the IEEE/CVF Conference on Computer Vision and Pattern Recognition}, pages 4190--4200, 2023.

\bibitem[Wang et~al.(2023)Wang, Liu, Dong, Guo, Liu, Wang, and Yang]{wang2023robust}
Haiping Wang, Yuan Liu, Zhen Dong, Yulan Guo, Yu-Shen Liu, Wenping Wang, and Bisheng Yang.
\newblock Robust multiview point cloud registration with reliable pose graph initialization and history reweighting.
\newblock In \emph{Proceedings of the IEEE/CVF Conference on Computer Vision and Pattern Recognition}, pages 9506--9515, 2023.

\bibitem[Wang et~al.(2021{\natexlab{a}})Wang, Liu, Liu, Theobalt, Komura, and Wang]{wang2021neus}
Peng Wang, Lingjie Liu, Yuan Liu, Christian Theobalt, Taku Komura, and Wenping Wang.
\newblock Neus: Learning neural implicit surfaces by volume rendering for multi-view reconstruction.
\newblock \emph{arXiv preprint arXiv:2106.10689}, 2021{\natexlab{a}}.

\bibitem[Wang and Solomon(2019)]{wang2019deep}
Yue Wang and Justin~M Solomon.
\newblock Deep closest point: Learning representations for point cloud registration.
\newblock In \emph{Proceedings of the IEEE/CVF international conference on computer vision}, pages 3523--3532, 2019.

\bibitem[Wang et~al.(2004)Wang, Bovik, Sheikh, and Simoncelli]{wang2004image}
Zhou Wang, Alan~C Bovik, Hamid~R Sheikh, and Eero~P Simoncelli.
\newblock Image quality assessment: from error visibility to structural similarity.
\newblock \emph{IEEE transactions on image processing}, 13\penalty0 (4):\penalty0 600--612, 2004.

\bibitem[Wang et~al.(2021{\natexlab{b}})Wang, Wu, Xie, Chen, and Prisacariu]{wang2021nerf}
Zirui Wang, Shangzhe Wu, Weidi Xie, Min Chen, and Victor~Adrian Prisacariu.
\newblock Nerf--: Neural radiance fields without known camera parameters.
\newblock \emph{arXiv preprint arXiv:2102.07064}, 2021{\natexlab{b}}.

\bibitem[Wei et~al.(2021)Wei, Liu, Rao, Zhao, Lu, and Zhou]{wei2021nerfingmvs}
Yi~Wei, Shaohui Liu, Yongming Rao, Wang Zhao, Jiwen Lu, and Jie Zhou.
\newblock Nerfingmvs: Guided optimization of neural radiance fields for indoor multi-view stereo.
\newblock In \emph{Proceedings of the IEEE/CVF International Conference on Computer Vision}, pages 5610--5619, 2021.

\bibitem[Xu et~al.(2020)Xu, Li, Du, Zhang, and Liu]{xu2020reluplex}
Jin Xu, Zishan Li, Bowen Du, Miaomiao Zhang, and Jing Liu.
\newblock Reluplex made more practical: Leaky relu.
\newblock In \emph{2020 IEEE Symposium on Computers and communications (ISCC)}, pages 1--7. IEEE, 2020.

\bibitem[Xue et~al.(2024)Xue, Lu, and Chen]{xue2024hdmnet}
Weiyi Xue, Fan Lu, and Guang Chen.
\newblock Hdmnet: A hierarchical matching network with double attention for large-scale outdoor lidar point cloud registration.
\newblock In \emph{Proceedings of the IEEE/CVF Winter Conference on Applications of Computer Vision}, pages 3393--3403, 2024.

\bibitem[Yang et~al.(2023)Yang, Chen, Wang, Manivasagam, Ma, Yang, and Urtasun]{yang2023unisim}
Ze~Yang, Yun Chen, Jingkang Wang, Sivabalan Manivasagam, Wei-Chiu Ma, Anqi~Joyce Yang, and Raquel Urtasun.
\newblock Unisim: A neural closed-loop sensor simulator.
\newblock In \emph{Proceedings of the IEEE/CVF Conference on Computer Vision and Pattern Recognition}, pages 1389--1399, 2023.

\bibitem[Yen-Chen et~al.(2021)Yen-Chen, Florence, Barron, Rodriguez, Isola, and Lin]{yen2021inerf}
Lin Yen-Chen, Pete Florence, Jonathan~T Barron, Alberto Rodriguez, Phillip Isola, and Tsung-Yi Lin.
\newblock inerf: Inverting neural radiance fields for pose estimation.
\newblock In \emph{2021 IEEE/RSJ International Conference on Intelligent Robots and Systems (IROS)}, pages 1323--1330. IEEE, 2021.

\bibitem[Yu et~al.(2022)Yu, Peng, Niemeyer, Sattler, and Geiger]{yu2022monosdf}
Zehao Yu, Songyou Peng, Michael Niemeyer, Torsten Sattler, and Andreas Geiger.
\newblock Monosdf: Exploring monocular geometric cues for neural implicit surface reconstruction.
\newblock \emph{Advances in neural information processing systems}, 35:\penalty0 25018--25032, 2022.

\bibitem[Yuan et~al.(2020)Yuan, Eckart, Kim, Jampani, Fox, and Kautz]{yuan2020deepgmr}
Wentao Yuan, Benjamin Eckart, Kihwan Kim, Varun Jampani, Dieter Fox, and Jan Kautz.
\newblock Deepgmr: Learning latent gaussian mixture models for registration.
\newblock In \emph{Computer Vision--ECCV 2020: 16th European Conference, Glasgow, UK, August 23--28, 2020, Proceedings, Part V 16}, pages 733--750. Springer, 2020.

\bibitem[Zhang et~al.(2022)Zhang, Zhang, Fu, Zhou, Cai, Huang, Jia, Zhao, and Tang]{zhang2022ray}
Jian Zhang, Yuanqing Zhang, Huan Fu, Xiaowei Zhou, Bowen Cai, Jinchi Huang, Rongfei Jia, Binqiang Zhao, and Xing Tang.
\newblock Ray priors through reprojection: Improving neural radiance fields for novel view extrapolation.
\newblock In \emph{Proceedings of the IEEE/CVF Conference on Computer Vision and Pattern Recognition}, pages 18376--18386, 2022.

\bibitem[Zhang et~al.(2023)Zhang, Zhang, Kuang, and Zhang]{zhang2023nerflidar}
Junge Zhang, Feihu Zhang, Shaochen Kuang, and Li~Zhang.
\newblock Nerf-lidar: Generating realistic lidar point clouds with neural radiance fields.
\newblock \emph{arXiv preprint arXiv:2304.14811}, 2023.

\bibitem[Zhang et~al.(2024)Zhang, Zhang, Kuang, and Zhang]{zhang2024nerf}
Junge Zhang, Feihu Zhang, Shaochen Kuang, and Li~Zhang.
\newblock Nerf-lidar: Generating realistic lidar point clouds with neural radiance fields.
\newblock In \emph{Proceedings of the AAAI Conference on Artificial Intelligence}, volume~38, pages 7178--7186, 2024.

\bibitem[Zhang et~al.(2018)Zhang, Isola, Efros, Shechtman, and Wang]{zhang2018unreasonable}
Richard Zhang, Phillip Isola, Alexei~A Efros, Eli Shechtman, and Oliver Wang.
\newblock The unreasonable effectiveness of deep features as a perceptual metric.
\newblock In \emph{Proceedings of the IEEE conference on computer vision and pattern recognition}, pages 586--595, 2018.

\bibitem[Zheng et~al.(2024)Zheng, Lu, Xue, Chen, and Jiang]{zheng2024lidar4d}
Zehan Zheng, Fan Lu, Weiyi Xue, Guang Chen, and Changjun Jiang.
\newblock Lidar4d: Dynamic neural fields for novel space-time view lidar synthesis.
\newblock \emph{arXiv preprint arXiv:2404.02742}, 2024.

\end{thebibliography}
\newpage
\appendix
\section{Appendix}
\subsection{LiDAR Pose Representation of GeoNLF} The exponential map of $\xi \in \mathfrak{se}(3)$ can be expressed as follows:
\label{apd:1}
\begin{align}
\exp \left(\xi^{\wedge}\right) & =\sum_{n=0}^{\infty} \frac{1}{n!}\left[\begin{array}{ll}
\phi^{\wedge} & \rho \\
\textbf{0}^{T} & 0
\end{array}\right]^{n} \\
& =\textbf{I}+\sum_{n=1}^{\infty} \frac{1}{n!}\left[\begin{array}{ll}
\phi^{\wedge} & \rho \\
\textbf{0}^{T} & 0
\end{array}\right]^{n} \\
& =\left[\begin{array}{ll}
\textbf{I} & 0 \\
\textbf{0}^{T} & 1
\end{array}\right]+\sum_{n=1}^{\infty} \frac{1}{n!}\left[\begin{array}{cc}
\left(\theta a^{\wedge}\right)^{n} & \left(\theta a^{\wedge}\right)^{n-1} \rho \\
\textbf{0}^{T} & 0
\end{array}\right] \\
& =\left[\begin{array}{cc}
\sum_{n=0}^{\infty} \frac{1}{n!}\left(\theta a^{\wedge}\right)^{n} & \sum_{n=1}^{\infty} \frac{1}{n!}\left(\theta a^{\wedge}\right)^{n-1} \rho \\
\textbf{0}^{T} & 1
\end{array}\right] \\
& = \left[\begin{array}{cc}
\sum_{n=0}^{\infty} \frac{1}{n !}\left(\boldsymbol{\phi^{\wedge}}\right)^n & \sum_{n=0}^{\infty} \frac{1}{(n+1) !}\left(\boldsymbol{\phi}^{\wedge}\right)^n \boldsymbol{\rho} \\
\mathbf{0}^T & 1
\end{array}\right]\\
& = \left[\begin{array}{cc}
    \boldsymbol{R} & \textcolor{red}{\boldsymbol{J}}\boldsymbol{\rho} \\
    \mathbf{0}^T & 1
    \end{array}\right]
\end{align}
where $\boldsymbol{R}=\sum_{n=0}^{\infty} \frac{1}{n !}\left(\phi^{\wedge}\right)^n$ and $J=\sum_{n=0}^{\infty} \frac{1}{(n+1) !}\left(\phi^{\wedge}\right)^n$. We omit the coefficient $\boldsymbol{J}$ from the translation term. Consequently, rotation and translation are decoupled for optimization. Additionally, we observe that the Neural LiDAR Field exhibits heightened sensitivity to rotation. It is feasible to employ distinct optimization strategies for rotation and translation. For instance, optimizing rotation first and then optimizing translation parameters. In our experiments, simultaneous optimization of both rotation and translation yielded satisfactory results.

\subsection{Pose-free Methods} 
\label{apd:2}
\textbf{Implementation Details of GeoNLF and Pose-free Baselines.}
All experiments were conducted on a single NVIDIA GeForce RTX 3090 GPU. We follow LiDAR4D~\cite{zheng2024lidar4d} to use hybrid representation and uniformly sampled 768 points along each laser. The optimization of GeoNLF is implemented on Pytorch with Adam optimizer with a learning rate of $1\times10^{-2}$ decaying to $1\times10^{-4}$ for NeRF and $1\times10^{-3}$ decaying to $1\times10^{-5}$ for translation and $5\times10^{-3}$ decaying to $5\times10^{-5}$ for rotation. We choose $n = 4$ adjacent frames to construct the graph and the control factor of GRCD linearly increases from 0 to 0.5. Since previous pose-free methods are designed for images, we use Lidar-NeRF~\cite{tao2023lidarnerf} with sinusoidal encoding as the backbone for BARF (BARF-LN) and Lidar-NeRF with hash encoding for HASH (HASH-LN), and adopt the same learning rate of GeoNLF.

\begin{figure*}[h]
\centering
  \includegraphics[width=0.9\textwidth]{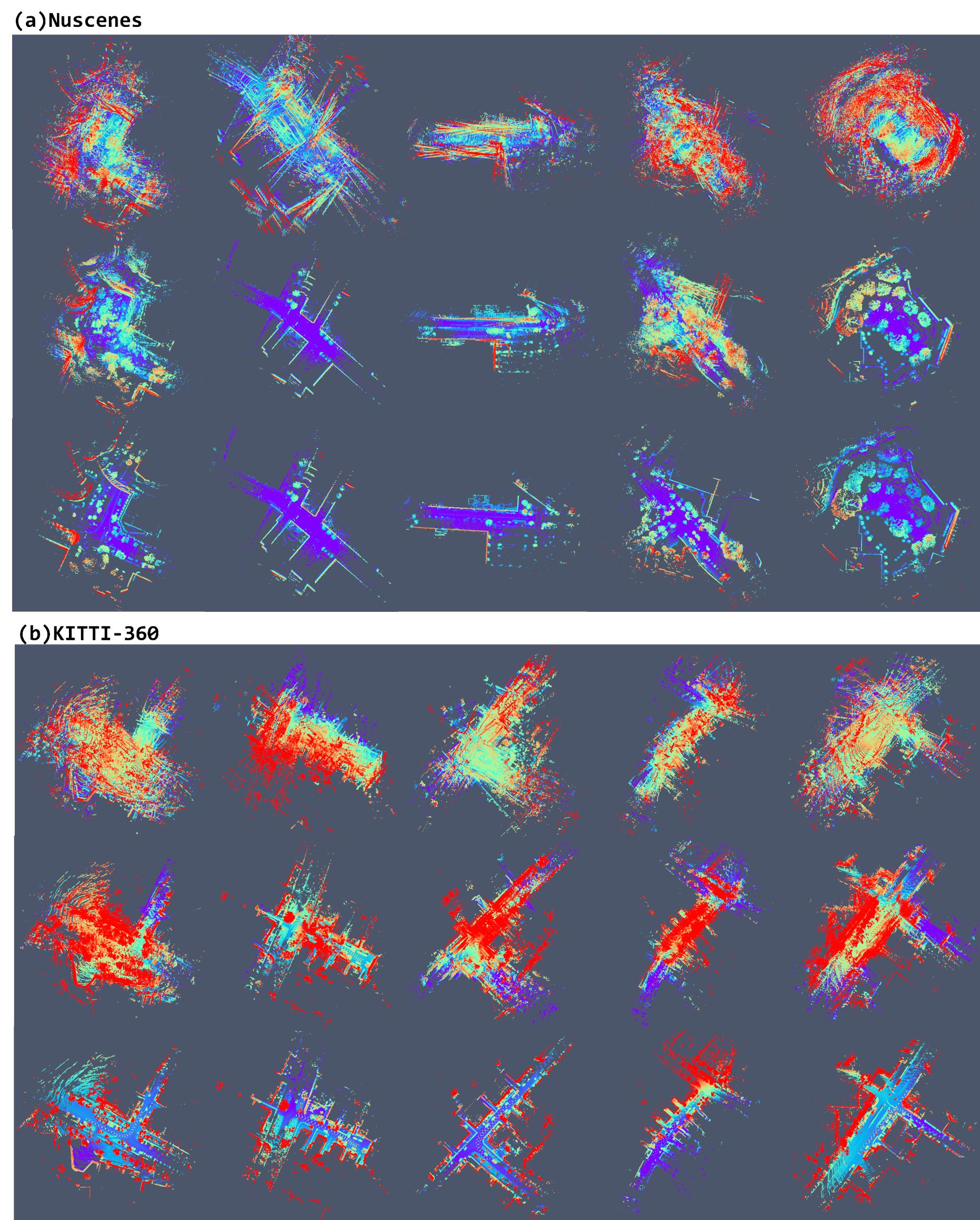}
  \caption{\textbf{Qualitative registration results of HASH-LN and GeoNLF on NuScenes and KITTI-360 dataset.}The first row contains original inputs, the second row shows the results of HASH-LN, and the third row displays the results of GeoNLF.}
  \label{fig:appendix0}
\end{figure*}

\textbf{Qualitative and Quantitative Registration Results of Pose-free Methods.}
Experimental data demonstrates that without the Geo-optimizer and SR training strategy, Pose-Free baseline HASH-LN/BARF-LN fails to effectively optimize translation error, leading to a decline in both $\mathrm{RPE_t},\mathrm{RPE_r}$ and $\mathrm{ATE}$. We present quantitative registration results of GeoNLF and HASH-LN for five sequences of NuScenes~\cite{caesar2020nuscenes} in \cref{exp:nvs}. Besides, we present qualitative registration results of 10 sequences in KITTI-360~\cite{liao2022kitti} and NuScenes~\cite{caesar2020nuscenes} dataset. As shown in \cref{fig:appendix0}, GeoNLF achieved favorable results across all 10 sequences, whereas HASL-LN succeeded only in a few sequences and fell into local optima in most cases.

\clearpage
\definecolor{best_result}{rgb}{0.96, 0.57, 0.58}
\definecolor{second_result}{rgb}{0.98, 0.78, 0.57}
\definecolor{third_result}{rgb}{1.0, 1.0, 0.56}
\begin{table*}[t]
\centering
\resizebox{\textwidth}{!}{
\renewcommand{\arraystretch}{1.25}
\begin{tabular}{ccccccccccccccc}
\hline \multirow[c]{2}{*}{ Method }  & \multicolumn{2}{c}{ Point Cloud } & \multicolumn{5}{c}{ Depth } & \multicolumn{5}{c}{ Intensity } \\
& CD$\downarrow$ & F-score$\uparrow$ & RMSE$\downarrow$ & MedAE$\downarrow$ & LPIPS$\downarrow$ & SSIM$\uparrow$ & PSNR$\uparrow$ & RMSE$\downarrow$ & MedAE$\downarrow$ & LPIPS$\downarrow$ & SSIM$\uparrow$ & PSNR$\uparrow$ \\
\hline
HASH-LN~\cite{heo2023robust,tao2023lidarnerf}\\
\hline
seq-1 & 0.8366 & 0.8108 & 8.7477 & 0.0294 & 0.1051 & 0.6714 & 19.2285 & 0.0502 & 0.0150 & 0.1031 & 0.6245 & 26.0173 \\	
seq-2 & 2.2041 & 0.6780 & 7.4556 & 0.1040 & 0.1977 & 0.4451 & 20.6918 & 0.0458 & 0.0134 & 0.1028 & 0.5329 & 26.8785 \\
seq-3 & 0.6510 & 0.8507 & 5.6282 & 0.0263 & 0.0975 & 0.7551 & 23.0687 & 0.0343 & 0.0091 & 0.0891 & 0.6929 & 29.2879 \\
seq-4 & 0.8330 & 0.7959 & 8.4737 & 0.0388 & 0.1263 & 0.6327 & 19.5087 & 0.0528 & 0.0169 & 0.1016 & 0.5871 & 25.5596 \\
seq-5 & 0.3210 & 0.8703 & 8.8714 & 0.0223 & 0.0685 & 0.7672 & 19.1347 & 0.0464 & 0.0130 & 0.0804 & 0.7021 & 26.6917 \\
\hline
\textbf{GeoNLF (ours)}\\
\hline
seq-1 & 0.2722 & 0.8638 & 5.7131 & 0.0275 & 0.0527 & 0.8192 & 22.9577 & 0.0396 & 0.0112 & 0.0762 & 0.7600 & 28.0745\\
seq-2 & 0.2957 & 0.8374 & 5.9331 & 0.0327 & 0.1361 & 0.6236 & 22.5965 & 0.0374 & 0.0091 & 0.0934 & 0.6760 & 28.5934\\
seq-3 & 0.1076 & 0.9015 & 3.8774 & 0.0222 & 0.0514 & 0.8590 & 26.2971 & 0.0267 & 0.0065 & 0.0567 & 0.7983 & 31.4740\\
seq-4 & 0.2561 & 0.8476 & 6.6385 & 0.0333 & 0.0752 & 0.7547 & 21.6418 & 0.0451 & 0.0134 & 0.0892 & 0.6986 & 26.9670\\
seq-5 & 0.2723 & 0.8731 & 6.9417 & 0.0246 & 0.0480 & 0.8168 & 21.2427 & 0.0402 & 0.0100 & 0.0714 & 0.7508 & 27.9300\\
\hline
\end{tabular}}
\caption{\textbf{Quantitative registration results of HASH-LN and GeoNLF on NuScenes dataset.}}
\label{exp:nvs}
\vspace{-.5cm}
\end{table*}

\clearpage
\subsection{Point Cloud Registration Methods} 
\label{apd:3}
\begin{figure*}[h]
\centering
  \includegraphics[width=0.95\textwidth]{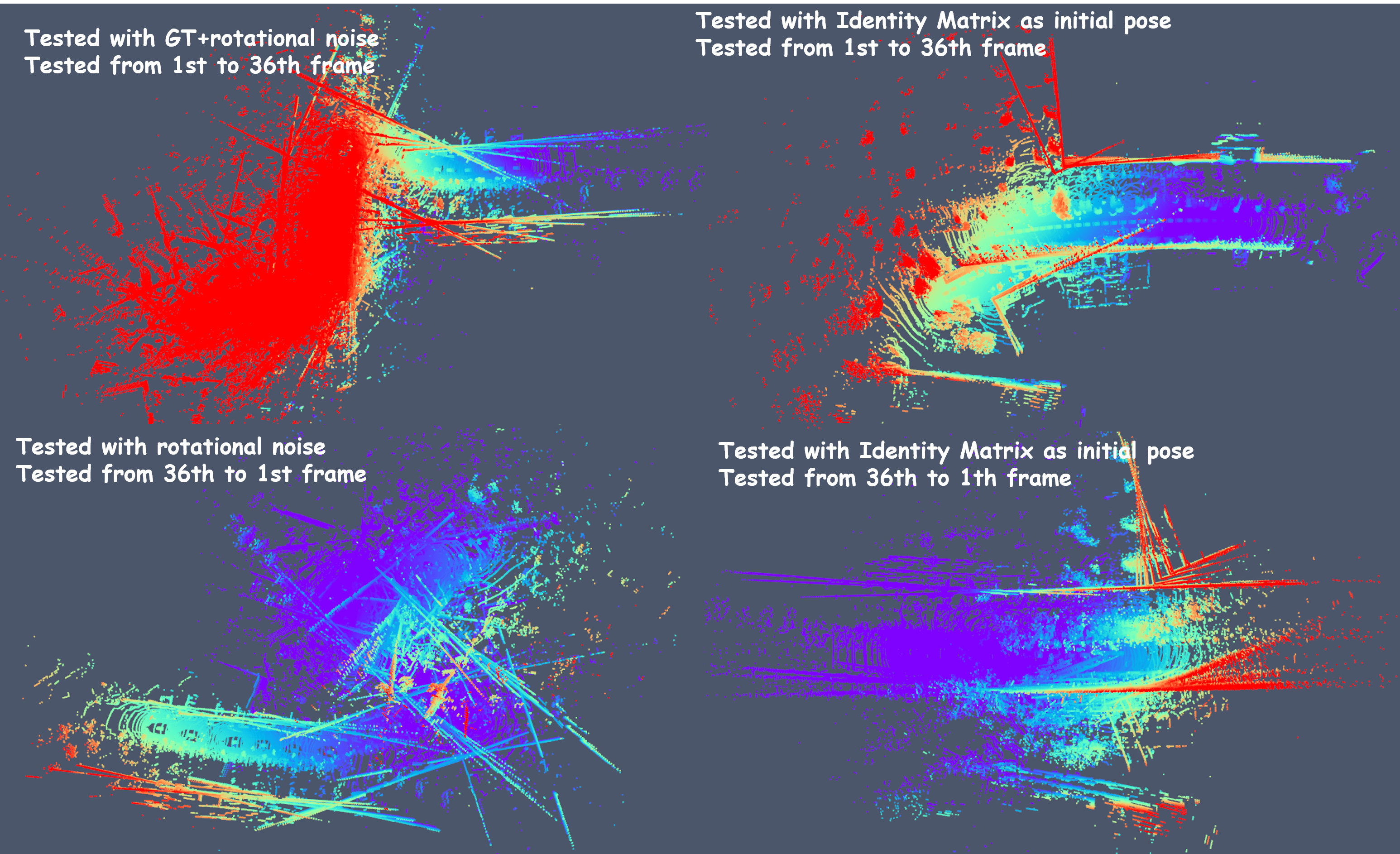}
  \caption{\textbf{Qualitative registration results of HRegNet on NuScenes (trained on KITTI).}}
  \label{fig:appendix1}
\end{figure*}

\begin{figure*}[h]
\centering
  \includegraphics[width=0.95\textwidth]{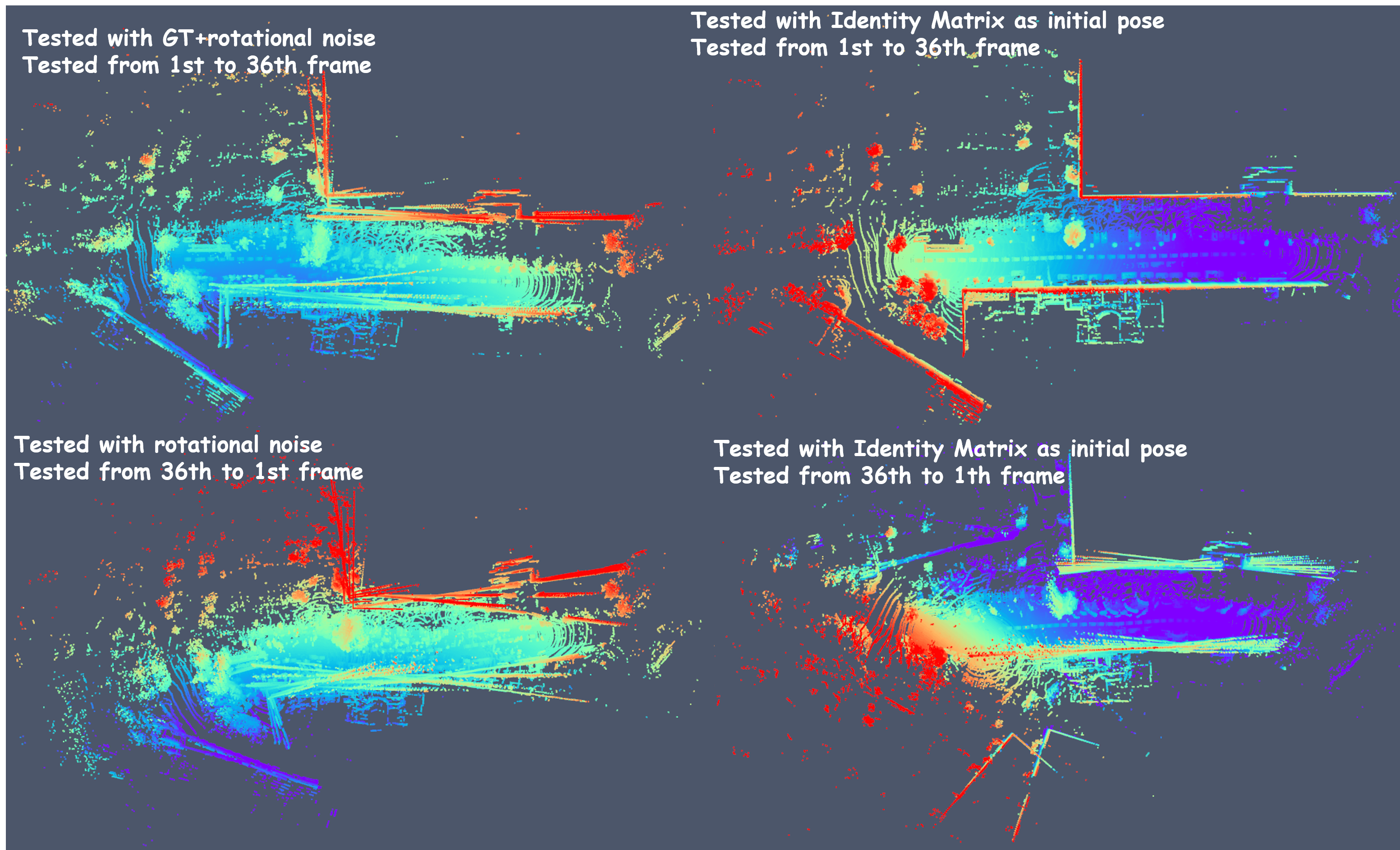}
  \caption{\textbf{Qualitative registration results of HRegNet on NuScenes (trained on NuScenes).}}
  \label{fig:appendix2}
\end{figure*}

All learning-based methods are trained on KITTI dataset~\cite{geiger2012we} according to the official training protocol, except for HRegNet~\cite{lu2021hregnet} due to its poor generalization. As shown in \cref{fig:appendix1}, HRegNet trained on KITTI performs poorly on NuScenes, rendering it unusable. We retrained HRegNet on NuScenes according to the official training protocol and tested it on our sequences in NuScenes. As shown in \cref{fig:appendix1} and \cref{fig:appendix2}: (1) Even when our sequences are part of the NuScenes training set, HRegNet produces poor results when random rotational noise is added. (2) HRegNet exhibits sensitivity to the temporal order of point cloud sequences. Registering frames sequentially from the 36th frame to earlier frames results in totally different outcomes compared to the reverse direction. 

\clearpage

\definecolor{best_result}{rgb}{0.96, 0.57, 0.58}
\definecolor{second_result}{rgb}{0.98, 0.78, 0.57}
\definecolor{third_result}{rgb}{1.0, 1.0, 0.56}
\begin{table*}[h]
\centering
\resizebox{\textwidth}{!}{
\renewcommand{\arraystretch}{1.25}
\begin{tabular}{ccccccccccccccc}
\hline 
\multirow[c]{2}{*}{Sequence} &\multicolumn{3}{c}{HRegNet}&\multicolumn{3}{c}{MICP} &\multicolumn{3}{c}{SGHR} &\multicolumn{3}{c}{ICP}\\
 &$\mathrm{RPE_t}$$\downarrow$ &$\mathrm{RPE_r}$$\downarrow$    &$\mathrm{ATE}$$\downarrow$
& $\mathrm{RPE_t}$$\downarrow$  & $\mathrm{RPE_r}$$\downarrow$  & $\mathrm{ATE}$$\downarrow$  
& $\mathrm{RPE_t}$$\downarrow$  & $\mathrm{RPE_r}$$\downarrow$  &$\mathrm{ATE}$$\downarrow$  
& $\mathrm{RPE_t}$$\downarrow$  & $\mathrm{RPE_r}$$\downarrow$  & $\mathrm{ATE}$$\downarrow$ \\
\hline 
seq-1    & 100.608 & 2.039 & 5.003     & 48.700 & 1.950 & 2.864     & 26.032 & 0.510 & 0.906    & 18.349 & 1.127 & 2.304    \\
seq-2    & 86.758 & 1.701 & 3.901     & 17.959 & 0.206 & 1.365     & 54.641 & 0.744 & 4.593    & 16.729 & 0.587 & 0.627   \\
seq-3    & 74.169 & 1.800 & 3.741     & 57.036 & 0.415 & 5.362     & 272.071 & 1.017 & 31.309    & 29.716 & 0.594 & 1.432   \\
seq-4    & 190.000 & 2.479 & 16.602     & 44.688 & 1.276 & 2.418    & 122.216 & 0.661 & 10.047    & 19.724 & 0.463 & 1.150   \\
seq-5    & 153.030 & 2.845 & 9.797     & 25.839 & 1.207 & 0.586     & 29.942 & 0.566 & 0.929    & 7.538 & 0.463 & 0.142    \\
\hline 
\end{tabular}}
\caption{\textbf{Quantitative Results of Registration on Nuscenes.} $\mathrm{RPE_t}$(cm), $\mathrm{RPE_r}$(deg), $\mathrm{ATE}$(m) }
\label{tab:pcr1}
\vspace{-.2cm}
\end{table*}
\begin{table*}[h]
\vspace{-.2cm}
\centering
\resizebox{\textwidth}{!}{
\renewcommand{\arraystretch}{1.25}
\begin{tabular}{ccccccccccccccc}
\hline 
\multirow[c]{2}{*}{Sequence} &\multicolumn{3}{c}{HRegNet}&\multicolumn{3}{c}{MICP} &\multicolumn{3}{c}{SGHR} &\multicolumn{3}{c}{ICP}\\
 &$\mathrm{RPE_t}$$\downarrow$ &$\mathrm{RPE_r}$$\downarrow$    &$\mathrm{ATE}$$\downarrow$
& $\mathrm{RPE_t}$$\downarrow$  & $\mathrm{RPE_r}$$\downarrow$  & $\mathrm{ATE}$$\downarrow$  
& $\mathrm{RPE_t}$$\downarrow$  & $\mathrm{RPE_r}$$\downarrow$  &$\mathrm{ATE}$$\downarrow$  
& $\mathrm{RPE_t}$$\downarrow$  & $\mathrm{RPE_r}$$\downarrow$  & $\mathrm{ATE}$$\downarrow$ \\
\hline 
seq-1    & 445.000 & 13.155 & 8.412     & 82.892 & 4.009 & 3.019     & 61.495 & 0.946 & 0.725    & 47.976 & 1.557 & 5.201    \\
seq-2    & 100.519 & 2.651 & 1.953     & 5.154 & 0.263 & 0.073     & 21.103 & 0.832 & 0.229    & 38.530 & 1.412 & 0.944   \\
seq-3    & 224.204 & 4.725 & 7.444     & 11.082 & 0.205 & 0.230     & 26.450 & 0.887 & 0.966    & 23.220 & 0.608 & 1.617   \\
seq-4    & 242.260 & 4.295 & 6.679     & 5.735 & 0.264 & 0.223     & 30.872 & 1.000 & 0.771    & 3.545 & 0.172 & 0.105   \\
seq-5    & 438.822 & 20.587 & 12.628     & 73.058 & 2.354 & 3.871     & 337.959 & 0.864 & 10.002    & 38.643 & 1.346 & 1.605    \\
\hline 
\end{tabular}}
\caption{\textbf{Quantitative Results of Registration on KITTI-360.} $\mathrm{RPE_t}$(cm), $\mathrm{RPE_r}$(deg), $\mathrm{ATE}$(m) }
\label{tab:pcr3}
\vspace{-.2cm}
\end{table*}

\begin{figure*}[h]
\centering
  \includegraphics[width=1\textwidth]{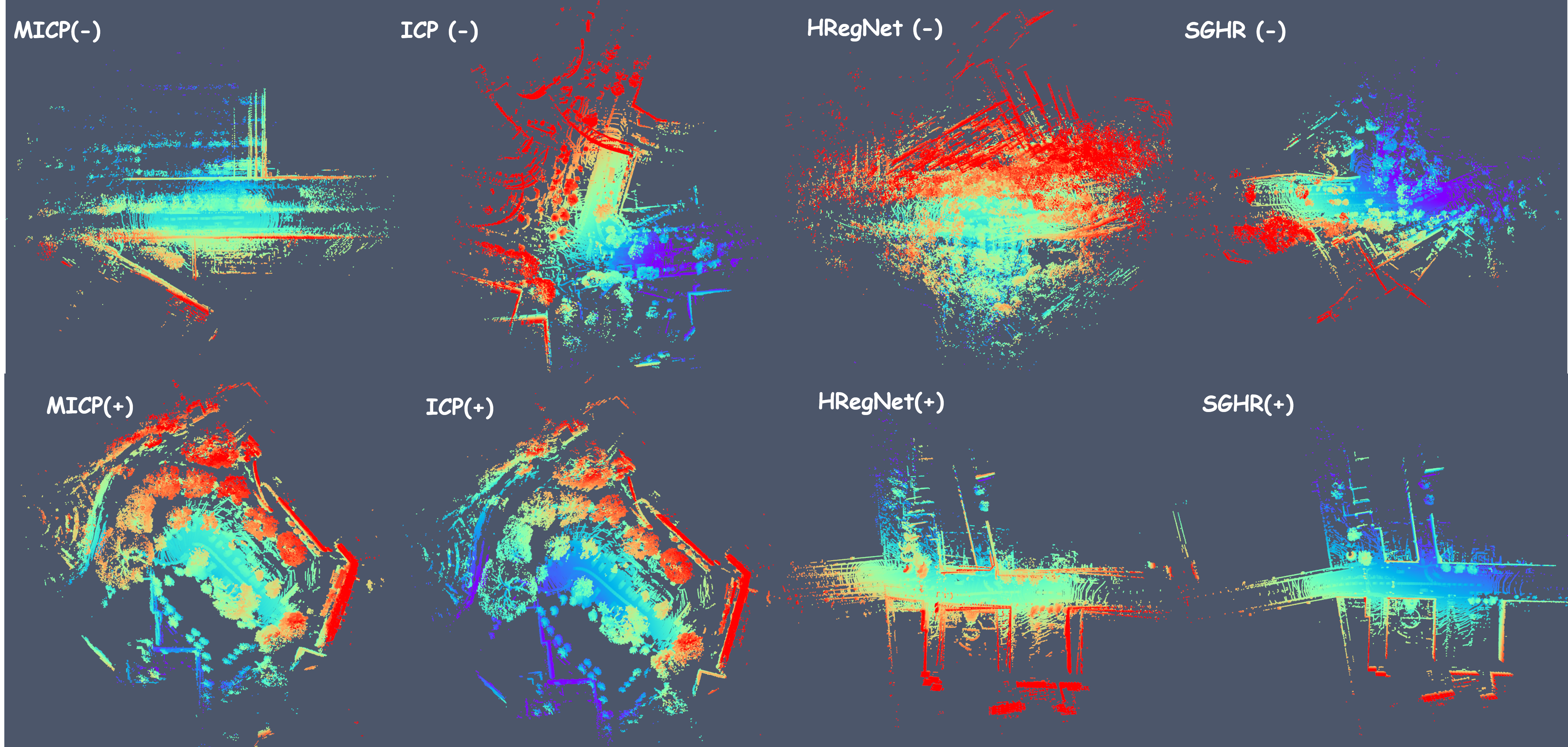}
  \caption{\textbf{Showcases of registration results of MICP, ICP, HRegNet, SGHR.} (-) represents the failure registration of the four methods, while (+) represents relatively successful registration.}
  \label{fig:appendix3}
\end{figure*} 

Generalization across different scenes is crucial for the effective use of PCR-assisted NeRF. Retraining on different datasets is not practically meaningful. However, all learning-based methods experience a significant performance drop when transferred to NuScenes. Moreover, point clouds from two temporally close frames are more likely to overlap and produce more reliable registration results. Under the strong assumption of a given point cloud sequence, multi-view registration methods, which align one frame with multiple others before refining the results, do not effectively utilize the sequence prior. Thus, the accuracy of MICP~\cite{choi2015robust} is inferior to pairwise methods in an odometry-like manner.

\textbf{Detailed Registration Results on different sequences of PCR Methods.} Except for GeoTrans~\cite{qin2022geometric}, HRegNet~\cite{lu2021hregnet}, ICP~\cite{besl1992method}, MICP~\cite{choi2015robust}, SGHR~\cite{wang2023robust} failed in many scenarios. We provide detailed registration results of HRegNet, ICP, MICP, SGHR for each sequence of in \cref{tab:pcr1} and \cref{tab:pcr3}. It can be observed that the registration methods are not sufficiently robust, often failing in certain scenarios. We illustrate successful registrations alongside failures for the four methods in \cref{fig:appendix3}.

\end{document}